\documentclass[lettersize,journal]{IEEEtran}
\usepackage{amsmath,amssymb,amsfonts}
\usepackage{algorithmic}
\usepackage{algorithm}
\usepackage{array}
\usepackage[caption=false,font=normalsize,labelfont=sf,textfont=sf]{subfig}
\usepackage{textcomp}
\usepackage{stfloats}
\usepackage{url}
\usepackage{verbatim}
\usepackage{graphicx}
\usepackage{cite}
\usepackage{makecell}
\usepackage{multirow}
\usepackage{xspace}
\usepackage[dvipsnames]{xcolor}
\usepackage{comment}
\usepackage[nolist,nohyperlinks]{acronym}
\usepackage[pagebackref=true,breaklinks=true,letterpaper=true,colorlinks,bookmarks=false]{hyperref}
\usepackage{orcidlink}

\usepackage{cleveref}
\usepackage{ifthen}
\usepackage{tikz}
\usetikzlibrary{calc,arrows,arrows.meta,decorations.pathreplacing}

\acrodef{GNN}{Graph Neural Network}
\acrodef{RAS}{Robotic-Assisted Surgery}
\acrodef{IIA-Net}{Instrument Interaction Aware Anticipation Network}
\acrodef{TCN}{Temporal Convolutional Network}
\acrodef{MAE}{Mean Absolute Error}
\acrodef{RSD}{Remaining Surgical Duration}
\acrodef{SOTA}{State-of-the-art}
\acrodef{SAM}{Segment Anything Model}
\usepackage{tabularx,colortbl}
\usepackage{xcolor} 
\usepackage{soul}
\sethlcolor{white}

\newcommand{\hlac}[1]{%
  \expandafter\hl\expandafter{\ac{#1}}%
}

\begin{document}
\title{Adaptive Graph Learning from Spatial Information for Surgical Workflow Anticipation}

\author{Francis Xiatian Zhang$^{\orcidlink{0000−0003−0228−6359}}$,~\IEEEmembership{Student Member,~IEEE}, Jingjing Deng$^{\orcidlink{0000-0001-9274-651X}}$,~\IEEEmembership{Member,~IEEE}, Robert Lieck$^{\orcidlink{0000-0003-0109-1285}}$, Hubert P. H. Shum$^{\orcidlink{0000-0001-5651-6039}\dag}$,~\IEEEmembership{Senior Member,~IEEE}
\thanks{F. X. Zhang, J. Deng, R. Lieck, H. P. H. Shum are with Durham University, UK. 
        (e-mail: \{xiatian.zhang, jinjing.deng, robert.lieck, hubert.shum\}@durham.ac.uk).}%
\thanks{$^{\dag}$Corresponding author: H. P. H. Shum}
}%

\markboth{Journal of \LaTeX\ Class Files,~Vol.~14, No.~8, August~2021}%
{Zhang \MakeLowercase{\textit{et al.}}: Adaptive Graph Learning from Spatial Information}

\IEEEoverridecommandlockouts
\IEEEpubid{\makebox[\columnwidth]{978-1-5386-5541-2/18/\$31.00~\copyright2018 IEEE \hfill} \hspace{\columnsep}\makebox[\columnwidth]{ }}
\maketitle
\IEEEpubidadjcol

\begin{abstract}
Surgical workflow anticipation is the task of predicting the timing of relevant surgical events from live video data, which is critical in \ac{RAS}. Accurate predictions require the use of spatial information to model surgical interactions. However, current methods focus solely on surgical instruments, assume static interactions between instruments, and only anticipate surgical events within a fixed time horizon.
To address these challenges, we propose an adaptive graph learning framework for surgical workflow anticipation based on a novel spatial representation, featuring three key innovations. First, we introduce a new representation of spatial information based on bounding boxes of surgical instruments and targets, including their detection confidence levels. These are trained on additional annotations we provide for two benchmark datasets. Second, we design an adaptive graph learning method to capture dynamic interactions. Third, we develop a multi-horizon objective that balances learning objectives for different time horizons, allowing for unconstrained predictions. Evaluations on two benchmarks reveal superior performance in short-to-mid-term anticipation, with an error reduction of approximately 3\% for surgical phase anticipation and 9\% for remaining surgical duration anticipation. These performance improvements demonstrate the effectiveness of our method and highlight its potential for enhancing preparation and coordination within the RAS team. This can improve surgical safety and the efficiency of operating room usage.
\end{abstract}


\section{Introduction}
\label{sec:introduction}

\IEEEPARstart{S}{urgical} workflow anticipation is the task of automatically predicting the timing of relevant surgical events from live video data, such as the remaining time before a surgical instrument is changed or the remaining duration of the entire surgery. The capability of this task facilitates efficient instrument preparation and the design of intelligent robotic assistance systems\mbox{\cite{rivoir2020rethinking}}. Additionally, it can enhance patient safety and facilitate communication in the operating room\mbox{\cite{maier2017surgical}}. Consequently, using machine learning models to improve surgical workflow anticipation has become an important research topic in surgical vision\mbox{\cite{aksamentov2017deep,twinanda2018rsdnet,rivoir2020rethinking,marafioti2021catanet}} and is critical for the efficiency and safety of Robotic-Assisted Surgery (RAS)\mbox{\cite{yuan2022anticipation}}.


\begin{figure}[t]
\centering
\includegraphics[scale = 0.50]{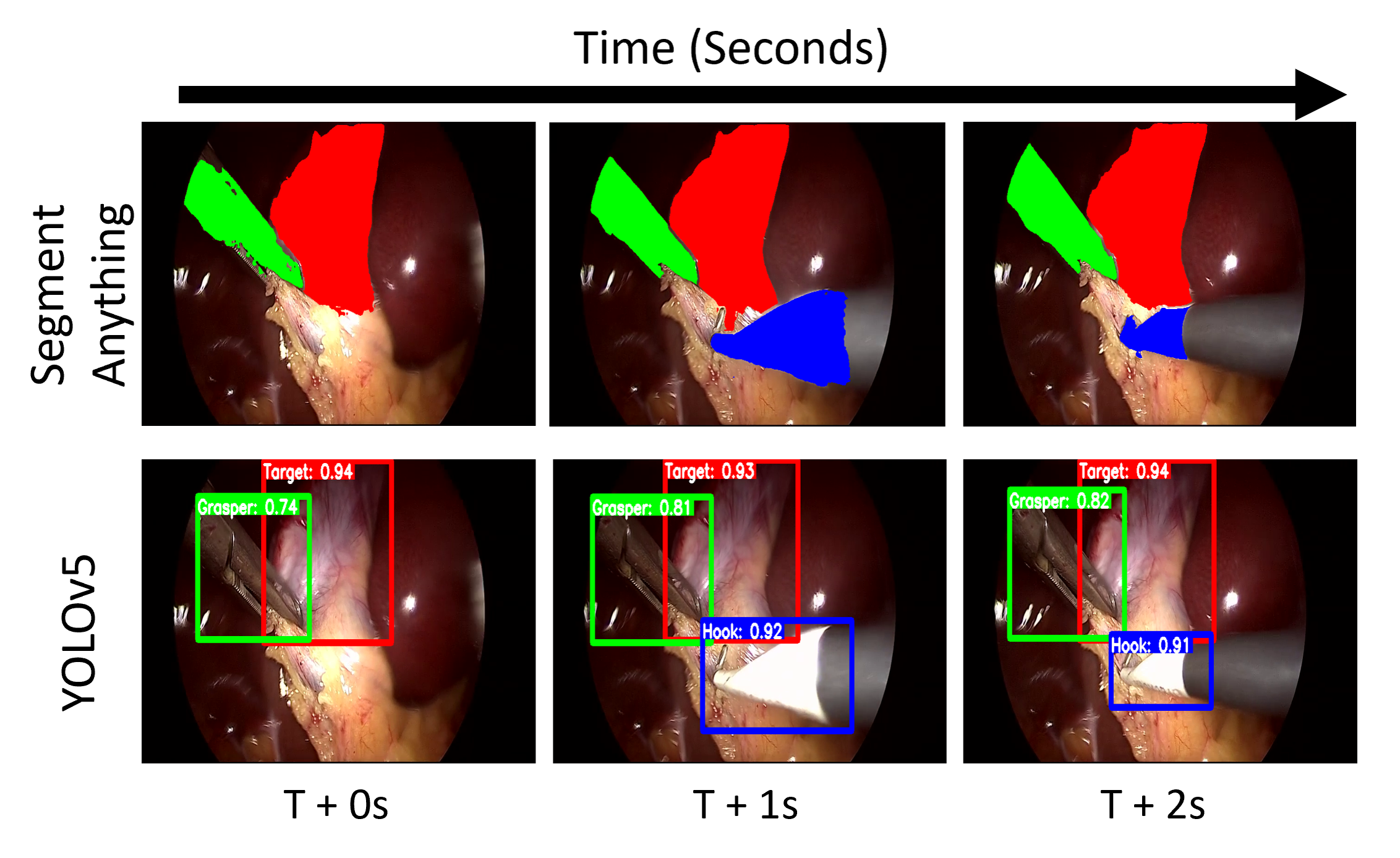}
\caption{Comparison of semantic segmentation (\textbf{Top}) with object detection (\textbf{Bottom}) across three consecutive seconds: We compare our trained YOLOv5 model with Segment Anything\mbox{\cite{kirillov2023segment}}, a state-of-the-art foundation model for segmentation. Segmentation masks significantly change across frames even when their positions remain static. In contrast, bounding boxes consistently provide a stable representation of both location and size.}
\label{fig:segmentation}
\end{figure}
To make accurate predictions, it is crucial to consider the interactions between surgical instruments and targets (\emph{e.g.,} a gripper fixating tissue) \cite{yuan2021surgical}, as these directly influence subsequent steps and surgical outcomes. \hl{This process requires integrating \textit{spatial information}\mbox{\cite{wang2023context}}, with a particular emphasis on tracking changes in location and size over time. Such tracking helps describe the rational movement and interaction relationships between surgical instruments and targets\mbox{\cite{li2021autonomous}}. Recent approaches\mbox{\cite{yuan2022anticipation,zhang2022towards}} utilize this spatial information, which is easily captured by existing object detection models and demonstrates robustness even during frequent partial occlusions in complex surgical environments\mbox{\cite{wang2022visual}} (as shown in Fig.~\mbox{\ref{fig:segmentation}}, where object detection could still detect robust spatial information for surgical instruments and targets during complex interactions).} However, these approaches\mbox{\cite{yuan2022anticipation,zhang2022towards}} only focus on location and size of instruments while ignoring surgical targets. Additionally, they assume interactions to be static over time\mbox{\cite{zhang2022towards}}, thereby overlooking the dynamic nature of the surgical workflow.

\hl{Anticipation models are generally trained and evaluated based on their prediction accuracy as measured by the Mean Absolute Error (MAE). In practice, immediately impending events are more important than those in the remote future. In the surgical workflow anticipation field, it is therefore common to define a so-called \textit{time horizon} $h$ (\emph{e.g.,} 2min/3min/5min)\mbox{\cite{rivoir2020rethinking,yuan2022anticipation}}, which sets a temporal threshold to differentiate between \textit{in-horizon} events ($t \leq h$) and \textit{out-of-horizon} events ($t > h$). These two types of events are evaluated differently\mbox{\cite{rivoir2020rethinking}}: in-horizon events occur within the given time horizon and should be predicted as accurately as possible, whereas out-of-horizon events occur beyond the given time horizon and should be largely ignored by setting the error to zero as long as the model predicts any time greater than $h$. This approach aims to better evaluate the accuracy and sensitivity of the anticipation model for the more relevant in-horizon events. However, previous works\mbox{\cite{rivoir2020rethinking,yuan2022anticipation}} typically use a fixed duration for $h$ (\emph{i.e.,} a \textit{fixed horizon}\mbox{\cite{wu2020learning}}). This approach limits the model’s ability to adapt to varying surgical durations because it cannot optimize for different time horizons within a single framework (\emph{i.e.,} \textit{multi-horizon} modeling\mbox{\cite{zhang2022towards}}). As the length of surgery can vary depending on the patient's condition\mbox{\cite{twinanda2018rsdnet}}, relying on a fixed time horizon restricts the model's generalizability across a broader patient population.}


In this paper, we propose a novel approach for surgical workflow anticipation that leverages spatial information and temporal dynamics. Our main contributions are (1)~a reliable spatial information representation based on bounding boxes and additional data annotations for instrument-target interactions, (2)~an approach for modeling spatial relations between instruments and targets over time using adaptive graphs, 
and (3)~an improved training strategy combining multiple fixed-horizon objectives into a single multi-horizon objective.
The identified challenges and our proposed solutions are now described in more detail.

\subsection{Challenges with Existing Approaches}
We identified three major challenges for employing spatial information for surgical workflow anticipation. 
First, existing methods provide an incomplete and unreliable representation of the surgical scene \cite{yuan2022anticipation,zhang2022towards}. Their representation is limited to surgical instruments, which ignores surgical targets as well as the uncertainty associated with the process of spatial information extraction. This lack of comprehensive information limits their ability to represent interactions effectively.

Second, prior methods use static graphs\mbox{\cite{sarikaya2020towards,zhang2022towards}} that cannot capture the dynamic interactions between instruments and surgical targets, which may vary significantly across the surgery. 
For example, in the cholecystectomy shown in Fig.~{\ref{fig:segmentation}}, the graspers are initially used to position the gallbladder for broad exposure. As the procedure progresses, the hook joins in to separate the gallbladder from the liver, necessitating a different interaction representation for the current frames. These changes in interaction cannot be captured by static graphs and require adaptive graphs.

Third, existing anticipation methods, including those based on spatial approaches\mbox{\cite{yuan2022anticipation,zhang2022towards}}, struggle with the diverse time span requirements of surgical anticipation. They typically employ a fixed time horizon in their model training and evaluation. When training, any prediction times exceeding the given time horizon are adjusted back to the fixed time horizon value. This approach limits their applicability to scenarios with a fixed horizon, while the time scales of surgical scenarios can vary significantly\mbox{\cite{twinanda2018rsdnet}}.

\subsection{Proposed Solutions}
To address these challenges, we present an adaptive graph learning framework for surgical workflow anticipation based on a novel spatial representation.
First, to represent the surgical scene more comprehensively, we propose a new spatial representation based on bounding boxes. This representation includes both surgical instruments and targets, along with their detection confidence levels.  Due to the lack of annotations for surgical targets in popular benchmark datasets, namely the Cholec80 cholecystectomy video dataset\footnote{\url{http://camma.u-strasbg.fr/datasets}} \cite{twinanda2016endonet} and the Cataract101 cataract surgery video dataset\footnote{\url{http://ftp.itec.aau.at/datasets/ovid/cat-101/}} \cite{SchoeffmannTSMP18}, we provide additional annotations for surgical targets. These annotations enable us to train object detection models for both surgical instruments and targets.

Second, we introduce an adaptive graph learning approach to promote a dynamic understanding of the surgical procedure. Unlike prior static methods for graph selection \cite{zhang2022towards}, our method dynamically selects suitable candidate graphs for each frame in the video thus adapting the graph representation over time according to the current surgical situation.

Third, to meet the diverse time horizon requirements of complex surgical settings, we introduce a multi-horizon objective that combines the loss functions for different time horizons using learnable weights. This optimizes a generalized prediction across different horizons without manual weight adjustment.

\subsection{Experiments}
Comprehensive experiments on two benchmark datasets indicate that our method outperforms existing surgical anticipation methods. To promote a fair comparison with previous methods, we employed the commonly used variants of Mean Absolute Error (MAE) as an evaluation metric, where a higher MAE implies inaccurate predictions that may delay the timely handling of relevant events and adversely affect patient outcomes\mbox{\cite{rivoir2020rethinking}}.
Specifically, compared to SOTA method proposed by Yuan et al.\mbox{\cite{yuan2022anticipation}}, our approach achieves a $\sim$3\% reduction in MAE for predicting the beginning of the next surgical phase (\textit{surgical phase anticipation}) and a $\sim$9\% reduction for predicting the end of the surgery (\textit{remaining surgery duration anticipation}) compared to existing benchmarks.
These MAE reductions demonstrate the ability of our framework to closely anticipate the actual surgical workflow, offering more precise assistance. This highlights the potential of our method for enhancing preparation and coordination within the RAS team, which helps improve surgical safety and the efficiency of operating room usage. Furthermore, our qualitative robustness analysis showcases that our bounding-box representation remains stable in the presence of visual artifacts such as motion blurring and illumination variations, making our method more robust than those based on raw pixel-level information.

The source code and our additional dataset annotations are available on GitHub.\footnote{\url{https://github.com/XiatianZhang/AdaAnticipation}} Our main contributions are the following:
\begin{enumerate}
\item We propose a new representation based on bounding boxes to extract spatial information about both instruments and surgical targets, along with their detection confidence levels. To train our object detection models and address the current gap in surgical target detection, we provide additional object annotations in two benchmark datasets.

\item We introduce an adaptive graph learning approach to dynamically select a number of candidate graphs for the current time frame that represent interactions among surgical instruments and targets. With graph convolution and temporal convolution, this allows for leveraging dynamic spatial information to improve our predictions in complex surgical settings.

\item We design a multi-horizon training strategy that incorporates loss terms for multiple time horizons, thus eliminating the need for choosing a single, fixed horizon. Our multi-horizon objective allows for automatically balancing the terms for different horizons, thereby making our approach more flexible and broadly applicable.
\end{enumerate}

\section{Related Work}
\subsection{Spatial Information Application for Surgical Scenarios}
Spatial information provides a structured framework for understanding complex scenarios and representing surgical environments effectively \cite{wang2023context,bronstein2017geometric}. Previous methods have utilized 3D models, such as meshes of surgical targets, to spatially represent these scenarios \cite{qian2019review}, which offers visualization guidance during surgery. These approaches heavily rely on the registration between 3D models and 2D surgical videos. This reliance can limit their utility in complex surgical environments, especially when occlusions occur \cite{qian2019review}.
Recent works fuse spatial information with visual information in deep learning  \cite{yuan2021surgical,yuan2022anticipation}. This fusion provides a more flexible representation and promotes automatic surgical assistance. Nevertheless, their complex designs for integrating spatial information extraction with other feature extraction methods often undermine the applicability of these approaches  \cite{yuan2022anticipation}.

To introduce a more structured understanding of surgical scenarios, recent surgical workflow analysis methods have adopted spatial information as their primary input. These frameworks utilize spatial information extracted from video as the input for their models. Currently, various types of spatial information are used in surgical scenarios, including key points \cite{sarikaya2020towards}, movement trajectories \cite{soleymani2022surgical}, and bounding boxes (\emph{i.e.,} location and size) \cite{zhang2022towards,lam2022deep,liu2022towards,zhang2023laparoscopic}. Nevertheless, their focus has been solely on the spatial details of instruments. This ignores the vital movements of surgical targets and results in an incomplete representation of surgical interactions. Additionally, they do not account for the potential uncertainty in their detection systems \cite{redmon2016you}. This ignorance often leads to reliance on unreliable information. To address these issues, we propose a more comprehensive representation that employs the spatial information of both instruments and surgical targets, including their confidence levels.

\subsection{Graph Learning in Surgical Video Analysis}
Graph representation learning emerges as a promising approach for surgical video analysis by effectively constructing structured representations from complex data \cite{hamilton2020graph}, such as the interactions between instruments and surgical targets during surgery. This structured representation enables efficient analysis of the spatial and temporal relationships among instruments and surgical targets.

Numerous studies have integrated graph learning into surgical applications \cite{wang2019graph,liu2021prototypical,liu2021graph,sarikaya2020towards,islam2020learning,long2021relational,kadkhodamohammadi2022patg,zhang2022towards}. These studies fall into two main categories: surgical scene understanding and surgical workflow understanding. In surgical scene understanding \cite{wang2019graph,liu2021prototypical,liu2021graph}, graph representations are constructed from extracted visual feature maps in an abstract manner. 
In surgical workflow understanding, graphs model the spatial perspective of surgical workflow and transform surgical scenarios into high-level semantic representations. Examples include representing tool joints or tool object detection outputs in graphs to conduct the surgical skills rating \cite{sarikaya2020towards} or workflow anticipation \cite{zhang2022towards} respectively.
While these methods have shown promising potential in surgical analysis, they often utilize a static graph to represent all interactions throughout surgery. This approach contradicts the dynamic nature of surgical interactions \cite{kennedy2020computer} and undermines the effectiveness of their representations. To overcome such limitations, our proposed adaptive graph learning dynamically selects candidate graphs representing different interactions.

\subsection{Surgical Workflow Anticipation}
Surgical workflow anticipation supports RAS by predicting the next likely step the surgeon will take \cite{yuan2022anticipation}. Unlike imminent anticipation tasks such as movement prediction \cite{shi2022recognition,kossowsky2022predicting,zhang2023laparoscopic}, surgical workflow anticipation focuses more on long-term event anticipation. Previous methods relied on pixel-level visual features from established visual feature extraction frameworks \cite{twinanda2018rsdnet,rivoir2020rethinking,marafioti2021catanet,yuan2021surgical,zhang2022towards,yuan2022anticipation}. For example, Rivor et al.\mbox{\cite{rivoir2020rethinking}} propose a framework that first extracts visual features from pixel-level information with 2D convolutions and then models these visual features with temporal modeling to output the final anticipation. These methods often fail to capture important semantic details, such as instrument interactions. This shortcoming significantly limits their anticipation performance because they lack a crucial understanding of the surgical progress through various surgical interactions.

Recent approaches employ spatial information to enhance the visual feature extraction backbone \cite{yuan2021surgical,yuan2022anticipation}. For example, Yuan et al.\mbox{\cite{yuan2022anticipation}} propose the {SOTA} method in surgical workflow anticipation for instruments and phases, Instrument Interaction Aware Anticipation Network (IIA-Net), which fuses non-visual features with visual features. It proposed an instrument interaction module into their feature extractor to reflect instrument interactions via spatial relations of instrument bounding boxes and semantic segmentation maps.
Additionally, Zhang et al.\mbox{\cite{zhang2022towards}} proposed a framework that uses spatial information as the primary input. They create a static graph representation based on extracted instrument bounding boxes, efficiently capturing and understanding instrument interactions based on their location and size. This structured approach improves the accuracy of predicting surgical events.

While achieving reasonable performance, recent methods struggle with the diverse time span requirements of surgical anticipation \cite{yuan2022anticipation,zhang2022towards}. They typically use a fixed time horizon for training and evaluation, which limits their ability to effectively address anticipation for different time horizons simultaneously. This approach is not effective because the suitable anticipation time horizon differs greatly among surgeries on different patients\mbox{\cite{twinanda2018rsdnet}}. To overcome this limitation, we introduce a multi-horizon training strategy into our learning framework.


\begin{figure}
    \centering
    \includegraphics[scale=0.40]{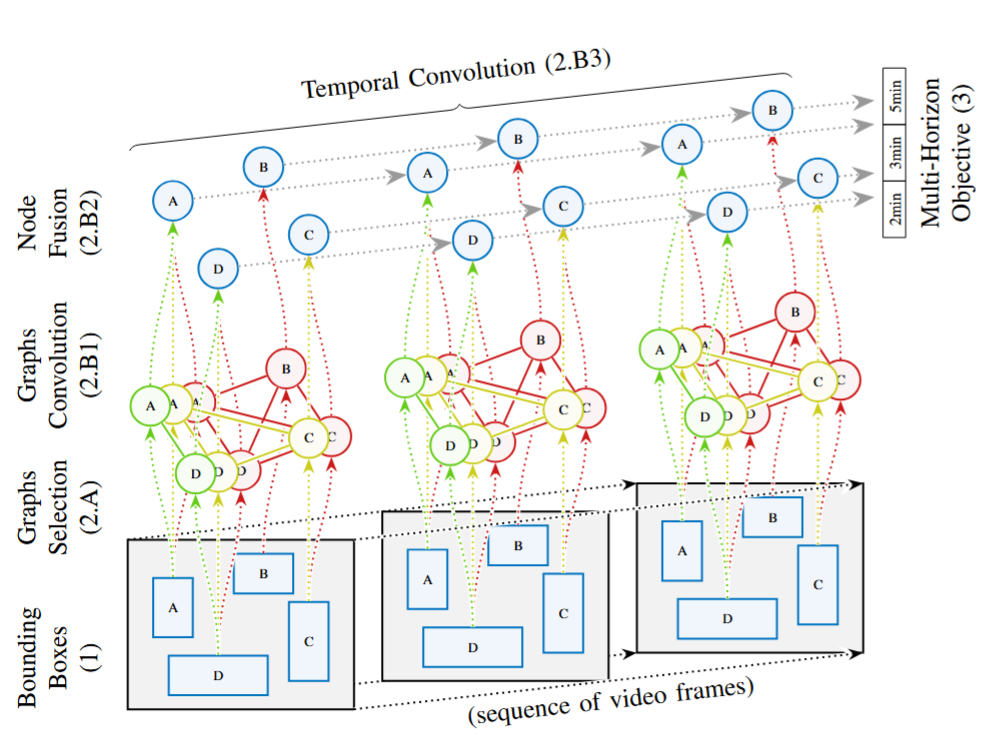}
    \caption{Overview of our method. Given a sequence of video frames as input, our model has three main processing stages (Sections {\mbox{\ref{sec:gi}}}--{\mbox{\ref{sec:aml}}}): (1):~From the raw frames, we extract bounding boxes of surgical instruments and targets. (2):~This information is further processed using adaptive graphs by (2.A)~selecting a number of candidate graphs (red, yellow, green), (2.B1)~using graph convolution to process the node features based on the graph's connectivity; (2.B2)~fusing nodes from the multiple candidate graphs, and (2.B3)~performing temporal convolution over the nodes from different video frames. (3):~The final node features are used to produce an unconstrained prediction of various surgical events trained using a multi-horizon objective.}
    \label{fig:method-overview}
\end{figure}

\begin{figure}
\centering
\includegraphics[scale = 0.25]{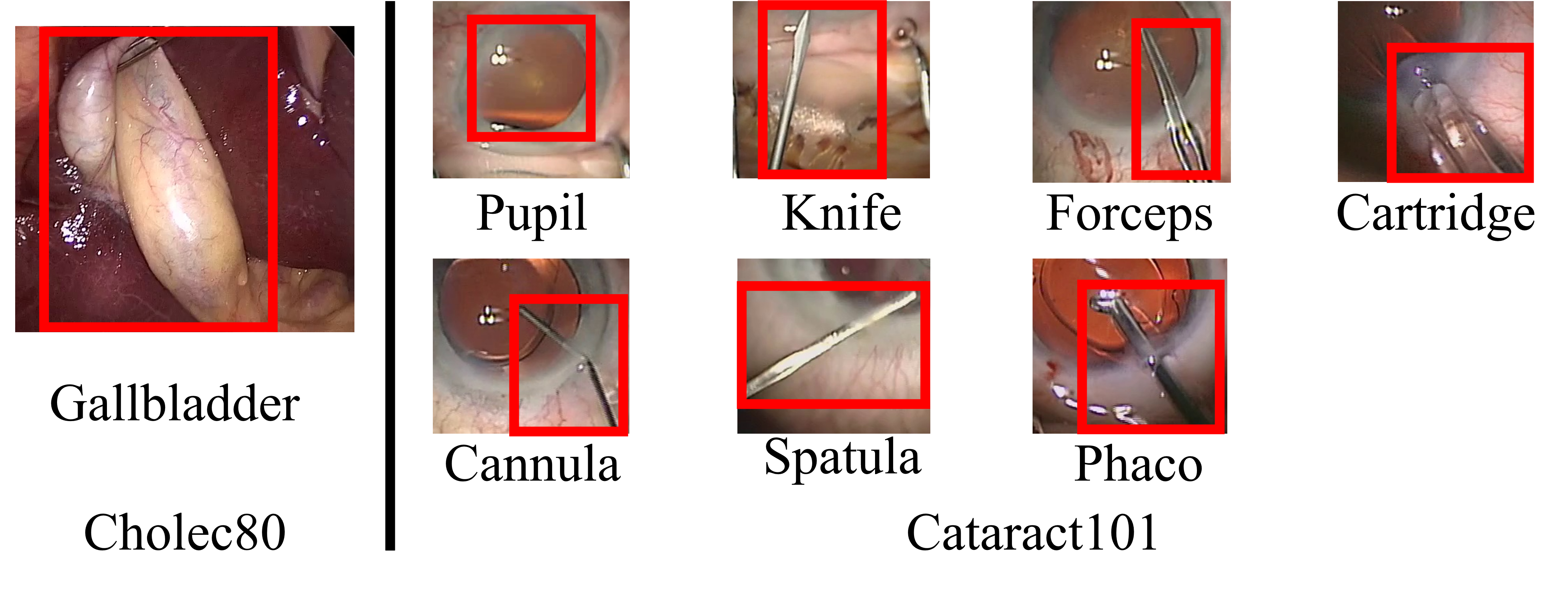} 
\caption{Additional annotation for existing datasets. For the Cholec80 dataset, we provide additional annotations that focus on surgical targets. For the Cataract101 dataset, we provide additional annotations that focus on surgical targets and surgical instruments.}
\label{fig:annot}
\end{figure}

\section{Method Overview}

Our method uses a series of raw video frames as input and predicts the timing of various surgical events as output. It consists of three main processing stages, depicted in Fig.~{\ref{fig:method-overview}} and described in more detail in Sections {\ref{sec:gi}}--{\ref{sec:aml}}.

\textbf{Spatial Information} (Section~{\ref{sec:gi}})\textbf{:} In the first stage, we extract bounding boxes from the raw video frames using YOLOv5\mbox{\cite{jocher2022ultralytics}}. By providing essential spatial information (\emph{i.e.,} location and size) with less complexity, bounding boxes offer a more stable spatial representation compared to pixel-level segmentation. At the same time, they provide all the necessary spatial information required for subsequent steps. To further improve the quality and robustness of the basic YOLOv5 model, we created additional annotations (Fig.~{\ref{fig:annot}})~and fine-tuned YOLOv5 on these datasets ({\ref{sec:1cFrhdQa}}). Moreover, we employ confidence estimates and a temporal attention mechanism to reweigh the extracted bounding boxes ({\ref{sec:deKki31d}}).

\textbf{Adaptive Graphs} (Section~{\ref{sec:agl}})\textbf{:} A core contribution of our work is the use of adaptive graphs to model the interaction between surgical instruments and targets over time. This involves two phases, first, the selection of candidate graphs based on both prior statistics about typical interactions as well as spatial information that is dynamically extracted from the bounding boxes ({\ref{sec:ags}}), and, second, further processing of the node features through graph convolutions, node fusion, and temporal convolutions ({\ref{sec:Gmeprka4}}).

\textbf{Multi-Horizon Objective} (Section~{\ref{sec:aml}})\textbf{:} It is common to train models for only predicting events within a fixed time horizon $h$ (2min/3min/5min) using dedicated loss functions. In contrast, our model is trained to predict the actual timing of events, independently of any artificially imposed time horizon constraints. When evaluating the model on existing fixed-horizon benchmarks, the model output is clipped to the respective horizon $h$. For training, we combine the loss functions for multiple horizons with a learnable weighting to a single multi-horizon loss. This allows our model to make unbounded predictions while still putting higher weight on the more relevant short-term predictions.


\section{Spatial Information Representation}
\label{sec:gi}
A major challenge in leveraging spatial information for surgical workflow anticipation\mbox{\cite{yuan2022anticipation,zhang2022towards}} is the lack of a stable representation of both instruments and surgical targets. This absence limits their effectiveness in representing critical interactions during surgery. Although previous efforts have utilized segmentations to represent surgical targets\mbox{\cite{yuan2022anticipation}}, this method primarily captures the shape of the surgical scene. Compared to object detection, segmentation often fails to directly describe the motion of instruments and surgical targets, which is crucial for understanding surgical workflows. Additionally, in the highly non-rigid surgical environment\mbox{\cite{kennedy2020computer}}, the shape of estimated segmentation masks frequently changes during interactions between instruments and targets. This variability often introduces noise into the training process, as demonstrated in the top part of Fig.~{\ref{fig:segmentation}}.

To address this challenge, our framework primarily employs bounding boxes to represent both instruments and surgical targets. It provides a stable spatial representation by focusing on key characteristics such as location and size, which remain consistent during interactions between instruments and targets, as illustrated in the bottom part of Fig. {\ref{fig:segmentation}}.

\subsection{Additional Data Annotation}
\label{sec:1cFrhdQa}
There are two significant issues with existing datasets. First, they do not include annotations for surgical targets\mbox{\cite{jin2018tool,SchoeffmannTSMP18}}. Second, most annotations are on selected high-quality images. This results in models being susceptible to image artifacts, such as reflection and motion blur, that can impair object detection performance.

To address these issues, we provide additional bounding box annotations on existing datasets for \textit{both} instruments \textit{and} surgical targets. Importantly, our annotations include a variety of image qualities without specifically selecting for high quality. This allows object detection models trained on our dataset to better handle the artifacts presented in real-world surgical settings.

Specifically, our additional annotations are based on two popular surgical video datasets: the Cholec80 cholecystectomy dataset \cite{twinanda2016endonet} and the Cataract101 cataract surgery dataset \cite{SchoeffmannTSMP18}. Figure {\ref{fig:annot} showcases examples of these additional annotations.}

For surgical target annotations, we define surgical targets as the tissue regions most frequently interacted with by instruments, identified based on our observation of the entire video for each surgical case. For the Cholec80 dataset, we provide additional annotations for 5,137 frames, focusing on surgical targets (\emph{i.e.,} the gallbladder and adjacent tissues targeted for removal). For the Cataract101 dataset, we provide additional annotations for 2,157 frames of surgical targets (\emph{i.e.,} the pupil).

For instrument annotations, our annotations leverage existing datasets. For Cholec80, we use the m2cai16-tool-locations dataset \cite{jin2018tool}. For the Cataract101 dataset, we provide additional annotations for 2,990 frames, employing a simplified instrument definition adapted from Fox et al. \cite{FoxTS20}.

We employ YOLOv5~\mbox{\cite{jocher2022ultralytics}} to extract spatial information (Section~{\ref{sec:deKki31d}}). YOLOv5 is known for its effectiveness in challenging conditions. This is due to its advanced data augmentation techniques that simulate visual distortions and its multiscale architecture capable of capturing objects of various sizes. We additionally fine-tune YOLOv5 on our new datasets to further enhance its object detection robustness. This enables us to accurately detect instruments or surgical targets despite environmental variability, facilitating effective spatial information extraction.

\subsection{Confidence Estimates and Temporal Attention}
\label{sec:deKki31d}
Previous anticipation methods \cite{yuan2022anticipation,zhang2022towards} that leverage spatial information (whether through bounding boxes or semantic segmentation) typically do not account for the uncertainty of their spatial information. This ignorance often leads to reliance on potentially inaccurate spatial information. Such overconfidence often leads to incorrect information being used for anticipation, particularly in complex surgical scenarios where the extraction of spatial information is noisy.

To address this issue, our approach includes bounding boxes and their detection confidence levels. This integration provides subsequent models with a quantifiable reference of potential uncertainty in the current spatial information extraction \cite{redmon2016you}. This method offers improved reliability over models that exclusively use segmentation or bounding boxes.

Specifically, we extract the bounding box features as:
\begin{equation}
b_{t} = \left\{ x_{t}, y_{t}, w_{t}, h_{t}, c_{t} \right\},
\end{equation}
where $x_{t}$ and $y_{t}$ denote the $x$ and $y$ coordinates of the center of a surgical element in frame $t$, $w_{t}$ and $h_{t}$ represent the detected width and height of each surgical element in frame $t$, and $c_{t}$ denotes the confidence level obtained from the object detection models. The observed sequence $\left\{ b_{0} \cdots b_{T} \right\}$, spanning from time point $0$ to the current observed time point $T$, is utilized to anticipate when $e$ occurs. Denoting the number of instruments and surgical targets in the surgery type to anticipate as $N$ and the feature number in the bounding box as $B$, the observed spatial information $b_{T}\in\mathbb{R}^{T\times N \times B}$. 

To further improve the robustness of our representation against potential inaccuracies, we adopt a learnable weighting for our representation to discern potential reliable frames through a temporal attention mechanism \cite{guo2022attention}. This mechanism assigns weights to frames based on object detection results and their associated confidence levels. The implementation details for obtaining the weighted spatial information $\hat{b}_{T} \in \mathbb{R}^{T\times N \times B}$ are as follows:
\begin{equation}
\begin{aligned}
Attn = \sigma(Conv1D(MaxPool(b_{T}))\\ + 
 Conv1D(AvgPool(b_{T})))
\end{aligned}
\end{equation}
\begin{equation}
\hat{b}_{T} = b_{T} \cdot Attn_{Temporal},
\label{eq:att_geo}
\end{equation}    
where $Attn \in \mathbb{R}^{T}$ denotes the temporal attention weight, $Conv1D$ denotes a causal 1D convolution. It facilitates online inference for each frame depending solely on current observed information \cite{yuan2022anticipation}. $\sigma$ denotes an activation function, $MaxPool$ and $AvgPool$ denote average and max pooling, respectively. $\hat{b}_{T}$ is then leveraged for subsequent adaptive graph learning.

\section{Adaptive Graph Learning}
\label{sec:agl}
A major challenge for previous anticipation methods\mbox{\cite{zhang2022towards}} employing spatial information is their reliance on a single static graph to describe the interaction relationships between instruments and surgical targets. This approach assumes fixed surgical interactions throughout the surgical procedure and fails to accurately capture the dynamic interactions between various instruments and surgical targets during surgery. Consequently, a single static graph often cannot offer an appropriate representation of interactions for different frames within a surgical video.

\begin{figure}
    \centering
    \includegraphics[scale = 0.17]{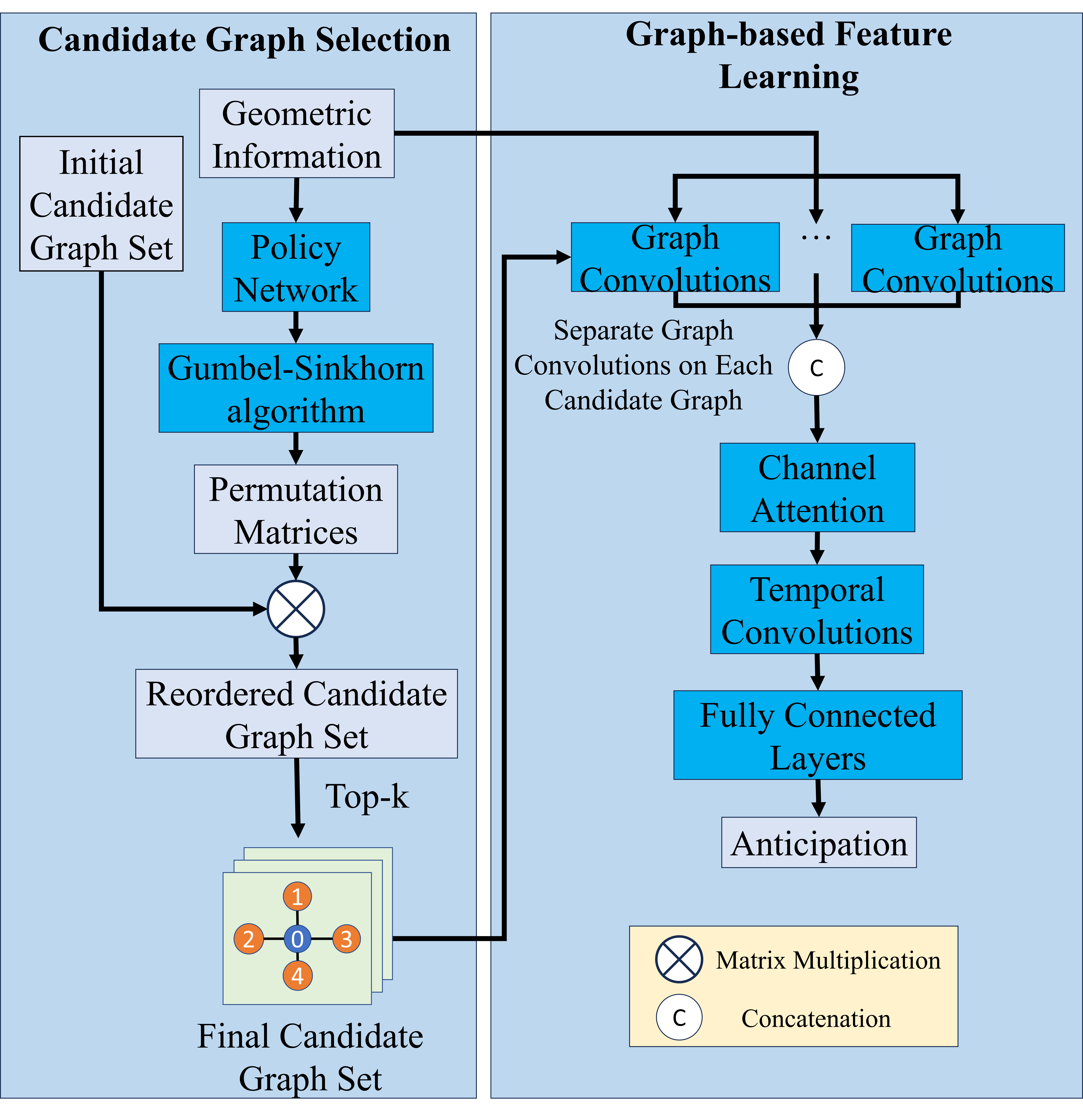}
    \caption{The architecture of our adaptive graph learning consists of two main components: \textbf{Left}: Candidate Graph Selection selects the suitable graph representations for each frame from the most common interactions observed in the training data. \textbf{Right}: Graph-based Feature Learning transforms spatial information into spatio-temporal features for anticipation based on the selected graphs for each frame, and then transforms the feature representation into the final anticipation output.}
    \label{fig:ADL}
\end{figure}

To address this challenge for dynamic surgical interactions, we design our anticipation feature learning to leverage adaptive graphs. These adaptive graphs automatically select different suitable representations for each frame, accurately reflecting the dynamic surgical interactions. Our method comprises two main steps: Candidate Graph Selection (Section~{\ref{sec:ags}}) which determines the graph representation policy for each frame, and Graph-based Feature Learning (Section~{\ref{sec:Gmeprka4}}) which transforms spatial information into spatio-temporal features for anticipation based on the selected graphs for each frame, and then transforms the feature representation into the final anticipation output. The procedure is illustrated in Fig.~{\ref{fig:ADL}}.
\subsection{Candidate Graph Selection}
\label{sec:ags}
\begin{figure}
\centering
\includegraphics[scale = 0.30]{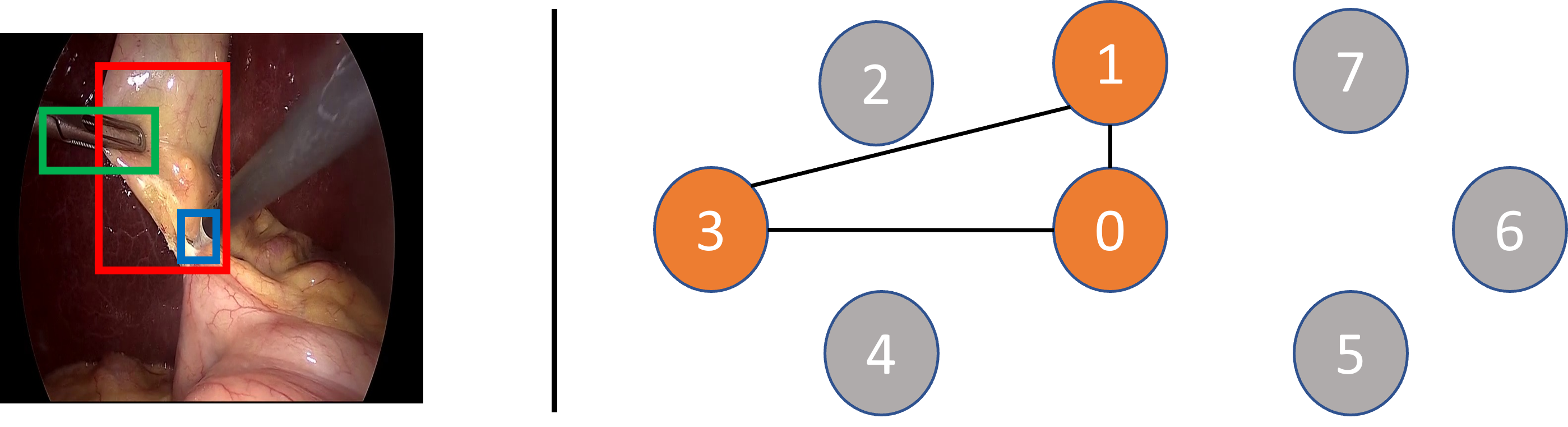}
\caption{Example of object detection results and graph representation from a cholecystectomy\mbox{\cite{twinanda2016endonet}}. \textbf{Left:} A frame showing the grasper and hook dissecting the tissue plane. \textbf{Right:} Fully connected candidate graph representing interactions among instruments and surgical targets. Gray nodes represent objects that do not appear in the frame. \textbf{Node legend:} 0: surgical target; 1: grasper; 2: bipolar; 3: hook; 4: scissors; 5: clipper; 6: irrigator; 7: specimen bag.
}
\label{fig:GCN}
\end{figure}
Two key issues arise when employing graph selections for surgical interaction analysis. First, the selected graphs should accurately represent surgical interactions\mbox{\cite{islam2020learning}}. Secondly, it is critical to design a graph selection process that is learnable to optimize it for the specific prediction task.

To address these issues, our candidate graph selection process involves two steps. First, before training, we generate an initial set of fully connected candidate graphs. This set includes the most frequently occurring combinations of instruments and surgical targets, according to object detection results from each frame in the training set. This approach identifies the most representative interactions within the dataset. Second, during training and inference, we employ a policy network enhanced by the Gumbel-Sinkhorn operation\mbox{\cite{mena2018learning}} to select the most relevant candidate graphs for each frame. The whole graph selection process is end-to-end differentiable, allowing us to optimize it for our anticipation tasks.

Specifically, in our first step, we construct a raw candidate graph set, denoted as $G$. This set comprises various graphs that encapsulate all possible combinations of instruments and surgical targets, which are identified from object detection results in each frame of the training data. Within these graphs, we define a node set $V$ with $N$ node where $N$ denotes the number of instruments or surgical targets in the current frame. Each node $v_{i}$ represents a bounding box of a surgical instrument or the surgical target and $N$ denotes the number of instruments or surgical targets in the current frame. The edge set $E$ encapsulates the interactions between these nodes, defined as $E = \{v_{i}v_{j} | i, j \in H, i \neq j\}$, where $H$ represents the set of all observed interactions based on the all possible combinations of instruments and surgical targets. Each graph in $G$ is associated with an $N \times N$ adjacency matrix $A$, where an entry $A^{ij} = 1$ signifies an interaction (\emph{i.e.,} an edge) between nodes $v_{i}$ and $v_{j}$. Figure \ref{fig:GCN} provides a visual example of object detection results and their corresponding graph representation.

To concentrate on the most representative interactions, we select $C$ graphs with the most common combinations according to their frequency of occurrence in the dataset. These form the initial candidate graph set $G_{C}  \in \mathbb{R}^{C}$. This set is then applied to the sequence of frames in our observed surgical videos, creating a corresponding sequence of initial candidate graph sets denoted as $G_{T} \in \mathbb{R}^{T \times C}$.

Then, in our second step, we derive a preliminary permutation likelihood matrix sequence $M_{T} \in \mathbb{R}^{T \times C \times C}$ from $\hat{b}_{T}$ via a policy network $Policy$, which is a $l_p$-layer causal \ac{TCN} \cite{yuan2022anticipation} with $l_p$ layers. This sequence guides the reordering of $G_{T}$:
\begin{equation}
M_{T} = Policy(\hat{b}_{T})
\end{equation}

To convert the likelihoods into a binary permutation matrix but still enabling gradient backpropagation for end-to-end learning, we utilize the Gumbel-Sinkhorn algorithm \cite{mena2018learning}, allowing for differentiable binarization of matrix $M_{t} \in \mathbb{R}^{C \times C}$ of each matrix from $M_{T}$. This approach first introduces a Gumbel noise term $GE_{t}$ to inject randomness into the permutation process, promoting that the model explores diverse permutations during training. Then, the Sinkhorn algorithm iteratively refines the matrix towards a doubly stochastic matrix that approximates binary permutation values:
\begin{equation}
\begin{aligned}
&M^{0}_{t} = exp(M_{t} + GE_{t}) \\
&M^{n}_{t} = \tau_{c}(\tau_{r}(M^{l-1}_{t})) \\
& \hat{M_{t}} = \textit{Softmax}\left(M^{n}_{t}/\tau\right),
\end{aligned}
\end{equation}
where $GE_{t} = -\log(-\log(U_{t}))$ and $U_{t}$ is sampled from a uniform $i.i.d$ distribution. $M^{n}_{t}$ denotes the permutation matrix at the $n$-th Sinkhorn iteration. This iterative process refines $M^{0}_{t}$ to approximate binary values by sequentially applying column-wise normalization $\tau_{c}$ and row-wise normalization $\tau_{r}$ to $P^{l-1}_{t}$. $\hat{M_{t}}\in \{0, 1\}^{C \times C}$ denotes the optimized permutation matrix of each frame and forms the permutation matrices $\hat{M}_{T}\in \{0, 1\}^{T \times C \times C}$ for observed frames. $\tau$ denotes the temperature parameter adjusting the sharpness of approximation.
Balancing computation cost and performance, we set the Sinkhorn iterations to 10. 

Finally, we reorder $G_{T}$ by multiplying it with the optimized permutation matrices $\hat{M}_{T}$. To focus on the most relevant candidate graphs for each frame, we leverage the top $k$ rows from the reordered matrices. This forms the final candidate graph set for each frame and a sequence of final candidate graph sets $\hat{G}_{T}  \in \mathbb{R}^{T \times k}$ for the observed video frames.

\subsection{Graph-based Feature Learning}
\label{sec:Gmeprka4}
A key limitation of existing graph-based feature learning in surgical workflow anticipation is the application of a fixed single graph and its graph convolutions to every frame in a surgery video\mbox{ \cite{zhang2022towards}}. This approach cannot accommodate the need for diverse graph convolutions to update graph node features based on varying surgical interactions because of the different graph representations selected across frames.

To overcome this limitation, we design a specific graph-based feature learning process that effectively incorporates selected candidate graphs. First, it employs independent graph convolutions for each selected candidate graph. This method promotes tailored updates to node features according to the distinct characteristics of different graphs. Second, we introduce a channel attention mechanism\mbox{\cite{guo2022attention}} to efficiently fuse node features across various graphs, providing the flexibility to integrate diverse graph combinations. Third, we incorporate dilated causal 2D convolutions to effectively aggregate current node features with those from previous frames. This embeds a broad temporal dimension into the feature learning to capture long-term temporal correlations.

In particular, first, each graph $\hat{G}_{T_k}$ from $\hat{G}_{T}$ is processed through 2-layers of spatial graph convolutions to encode spatial relationships and inject more semantic information into input spatial information \cite{kipf2017semi} where $\tilde{H}_{0} = \hat{b}_{T}$:
\begin{equation}
\tilde{H}_{l_g}^{(s)} = \Lambda_{k}^{-1/2} (A_{k} + I) \Lambda_{k}^{-1/2} \tilde{H}_{l_{g}-1}^{(k)} W_{l_g}^{(k)},
\end{equation}
where $\tilde{H}_{l_g}^{(k)} \in \mathbb{R}^{T \times C_{\tilde{H}} \times N}$ denotes the output features for the $k$-th selected graph after $l_g$ layers, $C_{\tilde{H}}$ is its channel number, $A_{k}$ denotes the adjacency matrix of the $k$-th selected graph, $\Lambda_{k}$ denotes the degree matrix, normalizing $\hat{G}_{T_k}$ for smooth information propagation. Notably, $W_{l_g}^{(k)}$ refers to the unique set of learnable convolution parameters allocated for the $k$-th graph, supporting node feature updates according to the specific graph structure. The collective feature set $\tilde{H}_{l_g} \in \mathbb{R}^{T \times C_{\tilde{H}} \times N \times k}$ concatenates the outputs across all selected graphs, creating a comprehensive representation that encapsulates diverse surgical interactions.

Second, our Squeeze-and-Excitation channel attention mechanism \cite{hu2018squeeze} facilitates adaptive graph fusion:
\begin{equation}
\begin{aligned}
&Attn_{g} = \sigma(W_{g}AvgPool(\tilde{H}_{l_g}))\\
&H' = Mean_{S}(\tilde{H}_{l_{g}} Attn_{g}),
\end{aligned}
\end{equation}
where $H' \in \mathbb{R}^{T \times C_{H'} \times N}$ represents the attention-weighted spatial graph features, $C_{H'}$ is its feature channel number, $Attn_{g} \in \mathbb{R}^{T \times S}$ is the attention matrix, $\sigma$ is the sigmoid activation function, $W_{g}$ represents the weight parameters, $AvgPool$ denotes the average pooling operation and $Mean_{S}$ denotes the mean operation along the optimal graph axis.

Third, our $l_t$-layer dilated causal 2D convolutions model temporal relationships across frames with more spatio-temporal interaction information where $\hat{H}_{0} = H'$:
\begin{equation}
\hat{H}_{l_{t}} = Conv2D_{l_t}(\hat{H}_{l_{t}-1} ),
\end{equation}
where $\hat{H}_{l_{t}} \in\mathbb{R}^{C_{\hat{H}} \times T \times  N} $ denotes the temporal feature output in $l_t$ layer, $C_{\hat{H}}$ is its channel number, $Conv2D_{l_t}$ denotes causal 2D convolutions that apply $2^{l_t-1}$ dilation along the time axis to broaden the receptive field in deeper layers, promoting ability to understand broader temporal contexts at a manageable computational cost \cite{yuan2022anticipation}. The causal design facilitates online inference for each frame \cite{yuan2022anticipation}.

Finally, anticipation outcomes $\hat{Y}_T \in \mathbb{R}^{T \times E}$ are generated from our feature $\hat{H}_{l_{t}}$ through a 2-layer fully connected neural network with activation functions.

\section{Multi-Horizon Objective}
\label{sec:aml}
\begin{figure}
    \centering
    \includegraphics[scale = 0.16]{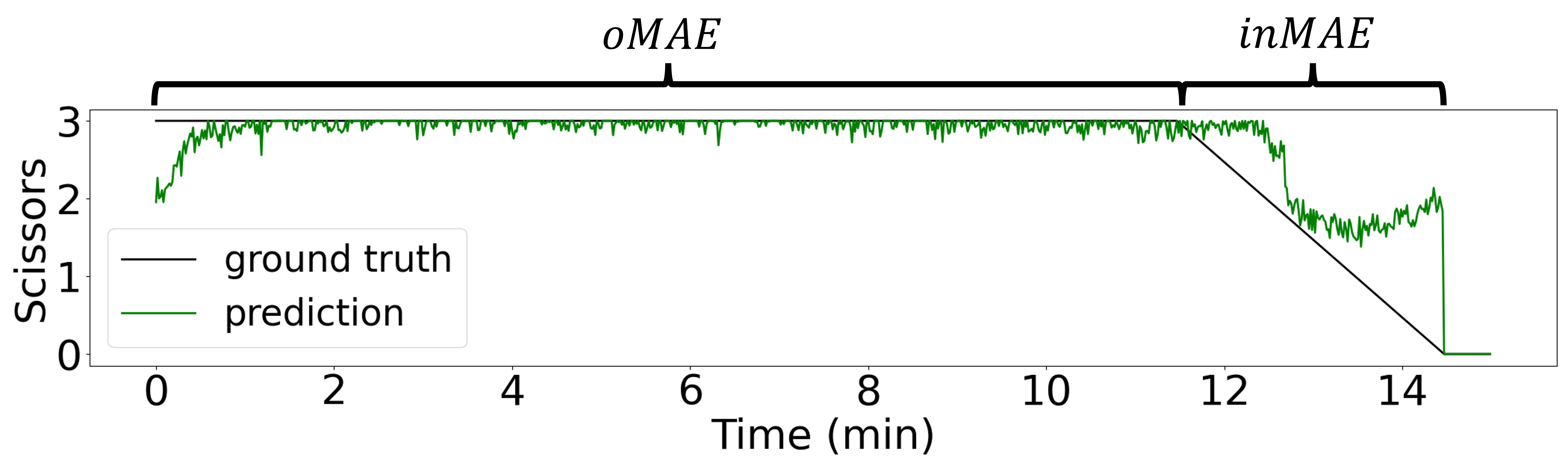}
    \caption{Example illustration of evaluating a model for a fixed time horizon of $h=3\mathrm{min}$. The relevant event (use of scissors) occurs at $t=14\mathrm{min}$. The ground truth is clipped to be between 0 and $h$. 
    }
    \label{fig:MAE}
\end{figure}
A major challenge in previous works\mbox{\cite{rivoir2020rethinking,zhang2022towards,yuan2022anticipation}} is achieving balanced anticipation performance for diverse time horizons. Previous works train and evaluate models using objectives for a fixed time horizon $h$, typically set at 2, 3, or 5 minutes. The model is supposed to predict the timing in the range of 0 to $h$, where 0 denotes that the event is currently occurring (\emph{e.g.}, a certain instrument is currently being used) and $h$ denotes that the (beginning of the) event lies $h$ or more in the future (this is also illustrated in Fig.~{\ref{fig:MAE}}). A problem with this setup is that setting the horizon $h$ is challenging. When $h$ is small, the model cannot anticipate any long-term events. Conversely, when the horizon is set large, the significant loss for long-term anticipation may compromise the optimization for short-term accuracy.

To address this challenge, we propose a multi-horizon objective training strategy that enables precise anticipation over different time horizons. Unlike previous methods with different outcomes of separate models for different $h$, our training and evaluation for different $h$ are based on one model and one outcome $\hat{Y}_T$, which increases the flexibility of our model. We introduce learnable variance representations\mbox{\cite{kendall2018multi}} to evenly distribute objectives across horizons, leading to a more balanced optimization process. This approach enables our training to adapt automatically to the optimal set of horizon weights for various surgical scenarios, enhancing the model's applicability and effectiveness.

In particular, learnable variance representations, denoted by $\lambda$, are adopted to adaptively normalize the loss for each training epoch for different tasks (\emph{i.e.,} horizons or different surgical events) \cite{kendall2018multi}. We employ two modified Mean Absolute Errors (MAEs) \cite{rivoir2020rethinking} to form our loss function, optimizing the anticipation for different parts of the ground truth: $inMAE$, where the ground truth is between $(0,h)$; $oMAE$, where the ground truth falls outside the $(0,h)$ range. Illustrations of these can be found in Fig.~\ref{fig:MAE}. If the event is more than $h$ in the future (out-of-horizon) the model is supposed to return a value of $h$, otherwise (in-horizon), it is supposed to predict the actual timing. For evaluation, the $MAE$ can be split into its in-horizon part ($inMAE$) and its out-of-horizon part ($oMAE$)\mbox{\cite{rivoir2020rethinking}}: $inMAE$ measures the accuracy of predicting imminent events, while $oMAE$ assesses the ability of a model to recognize that the event is not expected to occur in the near future. The multi-horizon loss $L$ is defined as:
\begin{equation}
\begin{aligned}
&\hat{\lambda}_{h} = SoftPlus(\lambda_{h}) \\
&L = oMAE_{H} + \sum\limits_{h}^{H} (\frac{inMAE_{h}}{2\hat{\lambda}_{h}} + \log(\hat{\lambda}_{h})),
\end{aligned}
\end{equation}
where $oMAE_{H}$ denotes the $oMAE$ for the largest horizon. For anticipation tasks without a fixed largest horizon, such as remaining surgery duration anticipation, this term is omitted. $inMAE_{h}$ and $\hat{\lambda}_{h}$ represent MAE and the learnable parameters for anticipation based on the specific horizon $h$, respectively. The $SoftPlus$ transformation converts $\hat{\lambda}_{h}$ to $\log(1+e^{\lambda_{h}})$, which ensures that the variance representation remains positive and aligns better with variance properties in multi-task learning scenarios. The terms $\log(\hat{\lambda}_{h})$ serve as a regularizer to prevent risks such as overfitting or learning instabilities from extremely large learnable variances \cite{kendall2018multi}.

At the beginning of training, each $\lambda_{h}$ is set to 1 to reflect an ideal situation where each horizon is balanced naturally \cite{kendall2018multi}. With this initialization, no specific horizon is favored over others, which prevents any pre-existing bias towards a specific horizon in our learning process for different surgeries.

\section{Experimental Setup}
\subsubsection{Data Split and Evaluation Procedures}
To demonstrate the effectiveness of our system, we test our method for two types of tasks: 1. Surgical instrument and phase anticipation: They predict the countdown time for when an instrument will be used and when a surgical phase will start, respectively. 2. Remaining surgical duration (RSD) anticipation: It predicts the countdown time for when a surgery will end.

For surgical instruments and phase anticipation, we primarily used the most common metrics from previous benchmarks\mbox{\cite{rivoir2020rethinking,yuan2022anticipation}}. Our main metric is $wMAE$, the average value of $inMAE$ and $oMAE$. As explained in Section \mbox{\ref{sec:aml}}, $inMAE$ measures the error for imminent in-horizon events that lie less than $h$ in the future, while $oMAE$ measures the error for out-of-horizon events that lie more that $h$ in the future. Thus, $wMAE$ provides an overall performance measure that balances in-horizon and out-of-horizon performance. Additionally, we use $eMAE$\mbox{\cite{yuan2021surgical,yuan2022anticipation}}, which measures the very-short-term prediction accuracy for events that lie less than $0.1\,h$ in the future (e.g.~$12\,$sec for a time horizon of $h=2\,$min). For all these metrics, lower values indicate better performance. A lower $wMAE$ indicates that our model consistently predicts both in-horizon and out-of-horizon events, which is crucial for its application across different types of surgeries. A lower $eMAE$ metric is particularly important for RAS\mbox{\cite{yuan2022anticipation}}, where the ability to make immediate adjustments to robotic operations during surgery is necessary.

For surgical instruments and phase anticipation, we conducted our comparison on the Cholec80 dataset \cite{twinanda2016endonet}. It includes 80 laparoscopic cholecystectomy videos, ranging from 15 to 90 minutes in duration. Adhering to the same data splitting and testing protocol in the benchmark \cite{twinanda2016endonet}, we processed videos at 1 frame per second (FPS), allocating 60 videos for training and 20 for testing. We conducted training four times, each with a different random seed. We reported the average performance over these training runs. Our task aimed to anticipate the occurrence of 5 surgical instruments and 6 surgical phases over horizons of 2, 3, or 5 minutes when they are not occurring \cite{yuan2022anticipation}. To facilitate meaningful outputs under multi-horizon training, we defined a function \( A(O) \) for the model outputs $O$:
\begin{equation}
A(O) = 
\begin{cases}
h & \text{if } O > h \\
O & \text{otherwise}
\end{cases}
\end{equation}

For RSD anticipation, we used the same metrics as previous benchmarks for evaluation \cite{marafioti2021catanet}: the MAE calculated from the start of 2 minutes remaining, 5 minutes remaining, and from the beginning for the entirety of the minutes remaining. Our comparison is conducted on the Cataract101 dataset \cite{SchoeffmannTSMP18}. It includes 101 cataract surgeries, with durations ranging from 2 to 20 minutes. Adhering to the same data splitting and testing protocol as established in the benchmark \cite{marafioti2021catanet}, we processed videos at 2.5 FPS and employed 6-fold cross-validation. The test results reflect the average performance across the 6 folds \cite{marafioti2021catanet}. The primary task focused on predicting the RSD for the entire duration of the procedure \cite{marafioti2021catanet}.

We employ online inference during both training and testing. For consistency with previous works, results on the Cholec80 dataset are reported solely with the mean error \cite{yuan2022anticipation}, while those on the Cataract101 dataset are presented with both mean error and standard deviation \cite{marafioti2021catanet}.
\subsubsection{Hyperparameters for Networks and Model Training}
The method was implemented in PyTorch 1.10 and trained on a Linux server with an Nvidia GeForce GTX 2080 Ti GPU. 

For our models applied to the Cholec80 and Cataract101 datasets, we customized the network hyperparameters to optimize performance for distinct tasks. Specifically, for Cholec80, we set both the policy network layers ($l_p$) and temporal convolution layers ($l_t$) to 8, expanding the receptive field to leverage more past information for accurate anticipation. For the Cataract101 dataset, specifically for RSD anticipation, we increased $l_p$ and $l_t$ to 11, enhancing the network capability for long-term understanding.

For both datasets, we set the number of initial graphs ($C$) and the number of selections ($k$) to 10 and 3, respectively. It is based on our statistical analysis showing that the top 10 graphs cover most interactions. $C = 10$ effectively captures the diversity of dynamic surgical interactions without overfitting to extremely rare interactions in the training data. Setting $k$ to 3 allows us to concentrate on the most relevant candidate graphs for each frame, promoting efficient graph selection.

The horizon sets for our multi-horizon objectives are \{2, 3, 5, 7\} minutes for the Cholec80 dataset and \{2, 5, $+\infty$\} (indicating an open-ended horizon) for the Cataract101 dataset. These intervals align with the range of single fixed training objectives found in previous works \cite{rivoir2020rethinking,yuan2022anticipation,marafioti2021catanet}. They are selected to cover a broad spectrum of short to long-term anticipations. During training, the ground truth values are clipped at the maximum horizon value to prevent overestimation. For model evaluation on benchmarks with predefined horizons, our output is adjusted to match the specified horizon $h$, facilitating compatibility with established evaluation standards.

We optimized additional hyperparameters using Bayesian hyper-parameter search on the Weights \& Biases platform \cite{wandb}. For the Cholec80 dataset, the optimization determined an epoch count of 100, a learning rate of 0.002, a weight decay of 0.00002, and a batch size of 2. Similarly, for Cataract101, the settings were an epoch count of 100, a learning rate of 0.0003, a weight decay of 0.00005, and a batch size of 4. Furthermore, to facilitate a fair comparison across benchmarks \cite{marafioti2021catanet}, we incorporated an auxiliary task during training to enhance the RSD anticipation.
\subsubsection{Benchmark Models}
For the instrument and phase anticipation, we compared our framework with the following recent methods: 1) TimeLSTM\mbox{\cite{aksamentov2017deep}}: A method proposed by Aksamentov et al.\mbox{\cite{aksamentov2017deep}}, which utilizes LSTM to model visual features extracted from videos and requires phase recognition labels during training. 2) RSDNet\mbox{\cite{twinanda2018rsdnet}}: A method by Twinanda et al.\mbox{\cite{twinanda2018rsdnet}}, similar to TimeLSTM\mbox{\cite{aksamentov2017deep}}, but does not require additional annotations. 3) TempAgg\mbox{\cite{sener2020temporal}}: A self-attention-based video anticipation method proposed by Sener et al.\mbox{\cite{sener2020temporal}}. 4) B-CNN-LSTM\mbox{\cite{rivoir2020rethinking}}: Introduced by Rivoir et al.\mbox{\cite{rivoir2020rethinking}}, this initial method for surgical instrument anticipation uses LSTM to model visual features from videos. 5) IIA-Net\mbox{\cite{yuan2022anticipation}}: The SOTA method proposed by Yuan et al.\mbox{\cite{yuan2022anticipation}}, which integrates visual and non-visual features for surgical workflow anticipation. 6) GCN-MSTCN\mbox{\cite{zhang2022towards}}: Proposed by Zhang et al.\mbox{\cite{zhang2022towards}}, it is an early approach that uses spatial information as the primary input.

For the \ac{RSD} anticipation, we compared our framework with the following recent methods: 1) TimeLSTM \cite{aksamentov2017deep}. 2) RSDNet \cite{twinanda2018rsdnet}. 3) TempAgg\mbox{\cite{sener2020temporal}}. 4) CataNet \cite{marafioti2021catanet}: The SOTA method by Marafioti et al.\mbox{\cite{marafioti2021catanet}}, which uses a similar design to TimeLSTM\mbox{\cite{aksamentov2017deep}} and requires phase recognition and surgeon experience labels during training. 5) GCN-MSTCN \cite{zhang2022towards}. We omitted other \ac{RSD} anticipation works \cite{wu2022nonlinear,wang2023real} from our comparison due to their different experimental settings.

We sourced performance data directly from \ac{SOTA} papers \cite{yuan2022anticipation,marafioti2021catanet}. For evaluating TimeLSTM \cite{aksamentov2017deep}, RSDNet \cite{twinanda2018rsdnet} on the instrument and phase anticipation task as well as TempAgg \cite{sener2020temporal} for both tasks, we retrained their model using our protocols and metrics. GCN-MSTCN \cite{zhang2022towards}, due to the different evaluation metrics used in their paper, we retrained their model using our protocols and metrics. 

\section{Experimental Results}
\begin{table*}[t]
\def\LEFTWIDTH{0.57}
\pgfmathsetmacro{\RIGHTWIDTH}{1-\LEFTWIDTH-0.01}
\pgfmathsetmacro{\LEFTWIDTH}{\LEFTWIDTH-0.01}
\def\COLSEP{\hspace*{3pt}}
\begin{minipage}[b]{\LEFTWIDTH\linewidth}
\scriptsize
\centering
\caption{$wMAE$ comparison on Cholec80 with \textbf{best} and \mbox{\underline{second best}} scores\\(AG: Adaptive Graph Learning, MHO: Multi-Horizon Objective).}
\begin{tabular}{@{\COLSEP}c@{\COLSEP}|@{\COLSEP}c@{\COLSEP}|@{\COLSEP}c@{\COLSEP}|@{\COLSEP}c@{\COLSEP}|@{\COLSEP}c@{\COLSEP}|@{\COLSEP}c@{\COLSEP}|@{\COLSEP}c@{\COLSEP}|@{\COLSEP}c@{\COLSEP}|@{\COLSEP}c@{\COLSEP}}
\hline
\multirow{2}{*}{\makecell{$wMAE$ \\ Comparison}}&\multicolumn{4}{@{\COLSEP}c@{\COLSEP}|@{\COLSEP}}{Instrument Anticipation}&\multicolumn{4}{@{\COLSEP}c@{\COLSEP}}{Phase Anticipation}\\
\cline{2-9}
 &2 min & 3 min & 5 min&$\text{Mean}_{2,3,5}$ &2 min & 3 min & 5 min&$\text{Mean}_{2,3,5}$ \\
\hline
TimeLSTM \cite{aksamentov2017deep} &0.51 &0.76 &1.32 &0.86 & 0.47 &0.64&1.07&0.72 \\
RSDNet \cite{twinanda2018rsdnet} & 0.48&0.73 &1.26&0.82 &0.43  &0.63&1.10&0.72 \\
TempAgg \cite{sener2020temporal} & 0.65 &0.92 &1.47 &1.01 & 0.38 &0.50&1.18&0.68 \\
B-CNN-LSTM\cite{rivoir2020rethinking}
& 0.43 &0.66 &1.09&0.73 & 0.39 &0.59 &0.85&0.61 \\
IIA-Net \cite{yuan2022anticipation} & \textbf{0.38} & \underline{0.58} & \textbf{0.92} & \textbf{0.63} & \underline{0.36} & \underline{0.49}&\textbf{0.68} & \textbf{0.51} \\
GCN-MSTCN \cite{zhang2022towards}& 0.48&0.72&1.21&0.80 &0.45 &0.67 &1.06&0.73 \\
\hline
Ours & \underline{0.39}& \textbf{0.57}&1.03 & \underline{0.66} & \textbf{0.35} & \textbf{0.47}&\underline{0.80}&\underline{0.54} \\
w/o AG&0.45 &0.70&1.33 &0.83 &0.45 &0.66 &1.16&0.76 \\
w/o MHO&0.43&0.61&\underline{1.00}&0.68 &0.37 &0.51 &\underline{0.80}&0.56 \\
\hline
\end{tabular}
\label{tab:wMAE}
\end{minipage}
\hfill
\begin{minipage}[b]{\RIGHTWIDTH\linewidth}
\scriptsize
\centering
\caption{$eMAE$ comparison with \textbf{best} and \mbox{\underline{second best}} scores.\\~}
\begin{tabular}{@{\COLSEP}c@{\COLSEP}|@{\COLSEP}c@{\COLSEP}|@{\COLSEP}c@{\COLSEP}|@{\COLSEP}c@{\COLSEP}|@{\COLSEP}c@{\COLSEP}|@{\COLSEP}c@{\COLSEP}|@{\COLSEP}c@{\COLSEP}}
\hline
\multirow{2}{*}{\makecell{$eMAE$ \\ Comparison}}&
\multicolumn{3}{@{\COLSEP}c@{\COLSEP}|@{\COLSEP}}{Instrument Anticipation}&\multicolumn{3}{@{\COLSEP}c@{\COLSEP}}{Phase Anticipation}\\
\cline{2-7}
&
 2 min & 3 min & 5 min&2 min & 3 min & 5 min \\
\hline
TimeLSTM \cite{aksamentov2017deep} &
1.56 &2.26 &3.62 & 1.29&1.73 &2.60 \\
RSDNet \cite{twinanda2018rsdnet} &
1.25&1.80 &2.61 & 1.15&1.55&2.40 \\
TempAgg \cite{sener2020temporal} &
1.27&1.35 &1.85 & 1.42 & 1.49&\underline{1.09} \\
B-CNN-LSTM\cite{rivoir2020rethinking} &
1.12	&1.65 &2.68 &1.02 &	1.47 &1.54\\
IIA-Net \cite{yuan2022anticipation} &
1.01 & 1.46& 2.14& 1.18 & 1.42 & \underline{1.09}\\
GCN-MSTCN \cite{zhang2022towards} &
\textbf{0.87}&1.26&1.90& \textbf{0.77}&\underline{1.06}&1.46\\
\hline
Ours & \underline{0.99}& \underline{1.21}&\underline{1.49}&0.95&\textbf{1.03 }&\textbf{1.06}\\
w/o AG&\textbf{0.87}&\textbf{1.06}&\textbf{1.34}&0.94&1.07&1.18\\
w/o MHO&\underline{0.99}&1.42&2.25&\underline{0.88}&1.17&1.63\\
\hline
\end{tabular}
\label{tab:eMAE}
\end{minipage}
\end{table*}

\begin{table}[t!]
\def\COLSEP{\hspace*{4.5pt}}
\scriptsize
\centering
\caption{MAE comparison for RSD Anticipation on Cataract101 with \textbf{best} and \underline{second best}: scores.}
\begin{tabular}{@{\COLSEP}c@{\COLSEP}|@{\COLSEP}c@{\COLSEP}|@{\COLSEP}c@{\COLSEP}|@{\COLSEP}c@{\COLSEP}|@{\COLSEP}c@{\COLSEP}}
\hline
Method & 2 min & 5 min &All & $\text{Mean}_{2, 5, \text{All}}$\\
\hline
TimeLSTM \cite{aksamentov2017deep} & 1.22$\pm$0.32 & 1.47$\pm$0.78 & 1.66$\pm$0.79 & 1.45 \\
RSDNet \cite{twinanda2018rsdnet} & 1.23$\pm$0.53 & 1.37$\pm$0.83 & 1.59$\pm$0.69 & 1.40 \\
TempAgg \cite{sener2020temporal} & 0.66$\pm$0.41 & 0.88$\pm$0.27 & 1.47$\pm$0.80 & 1.00\\
CataNet \cite{marafioti2021catanet} & \underline{0.35$\pm$0.20} & \underline{0.64$\pm$0.56} & \textbf{0.99$\pm$0.65} & \underline{0.66} \\
CataNet (Only RSD) \cite{marafioti2021catanet}& 
0.39$\pm$0.28 & 0.76$\pm$0.41 & 1.11$\pm$0.62 & 0.75 \\
GCN-MSTCN \cite{zhang2022towards} & 0.74$\pm$0.09 & 1.82$\pm$0.23 & 3.08$\pm$1.40 & 1.88 \\
\hline
Ours & \textbf{0.32$\pm$0.22} & \textbf{0.58}$\pm$\textbf{0.27} & \textbf{0.99$\pm$0.55} & \textbf{0.63} \\
\makecell{Ours (Only RSD) }& 0.36$\pm$0.26 & 0.65$\pm$0.31& \underline{1.00$\pm$0.52} & 0.67 \\
\hline
\end{tabular}
\label{tab:RSD}
\end{table}
\subsection{Comparison with Benchmarks}
Table \mbox{\ref{tab:wMAE}} shows the comparison of our proposed method with existing methods in surgical instruments and phase anticipation based on the widely used $wMAE$ metric. It is important to note that other methods, including the SOTA work IIA-Net in Table \mbox{\ref{tab:wMAE}}, trained individual models for each time horizon. In contrast, our approach trained a single model for all time horizons. Despite this, our method still demonstrates comparable or better performance for the 2-minute and 3-minute time horizons.

In instrument anticipation, we achieved comparable performance for the 2-minute time horizon and outperformed IIA-Net by 2\% for the 3-minute horizon. In phase anticipation, we outperformed IIA-Net by 3\% and 4\% for the 2-minute and 3-minute horizons, respectively. This superior performance demonstrates the generalizability of our proposed framework. These error reductions highlight the potential of our framework in assisting RAS by more accurately anticipating when surgical events will occur. For example, during a laparoscopic cholecystectomy, accurately predicting when the dissector will be needed allows the surgical team to prepare this instrument in advance\mbox{\cite{copenhaver2017improving}}. This reduces delays and improves the safety of the surgical procedure. Similarly, predicting the start of a surgical step, such as the dissection phase, enables the team to anticipate and coordinate their actions effectively\mbox{\cite{bogdanovic2015adaptive}}, enhancing overall surgical efficiency and safety.

For $eMAE$ (Table \ref{tab:eMAE}), our framework displayed improvements in both instrument and phase anticipation across all horizons. This indicates that our method is well-suited for rapid-response scenarios, providing accurate predictions even in very short time frames. Lower $eMAE$ scores are particularly important for evaluating very short-term anticipation, which is crucial for RAS where immediate adjustments are necessary. During RAS, the ability to quickly predict the need for specific instruments or actions significantly enhances the responsiveness of the surgical robots\mbox{\cite{rivoir2020rethinking}}. This further reduces response time and improves the coordination between the surgeon and the robots.

In \ac{RSD} anticipation (Table \ref{tab:RSD}), our model significantly outperformed both CataNet \cite{marafioti2021catanet} and GCN-MSTCN, with improvements of 9\% and 57\% for the 2-minute horizons, and 9\% and 68\% for the 5-minute horizons, respectively. The substantial accuracy reduction of GCN-MSTCN suggests the previous static graph method may not generalize well for various surgeries. Additionally, even when training our method without auxiliary labels, our method exhibited a significantly smaller performance reduction than CataNet \cite{marafioti2021catanet}, which demonstrates the generalization ability of our method. These error reductions for RSD highlight the potential of our proposed framework to not only benefit planning within a single surgery but also improve operation arrangements across different surgery patients. This advantage enhances the hospital's ability to manage operating rooms more effectively\mbox{\cite{marafioti2021catanet}}, which in turn provides more patients the opportunity to receive timely treatment.

A highlight of our approach is its use of graph-based data derived from bounding-box information, which emphasizes high-level interaction semantics between instruments and surgical targets. This focus allows us to provide a learning framework that is more cost-effective in terms of computational resources and implementation ease. Nevertheless, our design might occasionally overlook low-level details of the surgical context, such as texture information. Unlike other methods in Table \ref{tab:wMAE}, our model does not fine-tune for a single fixed time horizon. These factors could explain why our advantage in long-term anticipation is not as significant as it is for the 2-minute and 3-minute time horizons.

\begin{table*}[t]
\begin{minipage}{0.49\linewidth}
\scriptsize
\centering
\caption{$inMAE$ comparison with \textbf{best} and \mbox{\underline{second best}} scores.}
\begin{tabular}{c|c|c|c|c|c|c}
\hline
\multirow{2}{*}{\makecell{$inMAE$ \\ Comparison}}&\multicolumn{3}{c|}{Instrument Anticipation}&\multicolumn{3}{c}{Phase Anticipation}\\
\cline{2-7}
 &2 min & 3 min & 5 min &2 min & 3 min & 5 min \\
\hline
TimeLSTM \cite{aksamentov2017deep} &0.86 &1.24 &1.98 & 0.68 &0.96&1.54 \\
RSDNet \cite{twinanda2018rsdnet} & 0.73&1.05 &1.62 &0.65  &0.93&1.55 \\
TempAgg \cite{sener2020temporal} & 0.87 &1.20 &1.78 & 0.71 &0.95&1.40 \\
B-CNN-LSTM\cite{rivoir2020rethinking} & 0.77 &1.17&1.75 & 0.63 &0.86 &1.17 \\
IIA-Net \cite{yuan2022anticipation} & \underline{0.66} & 0.97 & \underline{1.40} & \underline{0.62} & \underline{0.81}&\underline{1.08} \\
GCN-MSTCN \cite{zhang2022towards}& \textbf{0.64}&\textbf{0.93}&1.43 &\textbf{0.61} &0.86 &1.28 \\
\hline
Ours & 0.72& \underline{0.96}&\textbf{1.33} & 0.63 & \textbf{0.77} &\textbf{1.06} \\
\hline
\end{tabular}
\label{tab:inMAE}
\end{minipage}
\begin{minipage}{0.49\linewidth}
\scriptsize
\centering
\caption{$oMAE$ comparison with \textbf{best} and \mbox{\underline{second best}} scores.}
\begin{tabular}{c|c|c|c|c|c|c}
\hline
\multirow{2}{*}{\makecell{$oMAE$ \\ Comparison}}&\multicolumn{3}{c|}{Instrument Anticipation}&\multicolumn{3}{c}{Phase Anticipation}\\
\cline{2-7}
 &2 min & 3 min & 5 min &2 min & 3 min & 5 min \\
\hline
TimeLSTM \cite{aksamentov2017deep} &0.16 &0.29 &\underline{0.67} & 0.22 &0.32&0.60 \\
RSDNet \cite{twinanda2018rsdnet} & 0.23&0.42 &0.90 &0.20  &0.33&0.65 \\
TempAgg \cite{sener2020temporal} & 0.43 &0.64 &1.16 & \textbf{0.06} &\textbf{0.06}&0.97 \\
B-CNN-LSTM\cite{rivoir2020rethinking} & \underline{0.08} &\textbf{0.15}&\textbf{0.44} & 0.15 &0.32 &\underline{0.52} \\
IIA-Net \cite{yuan2022anticipation} & 0.10 & \underline{0.19} & \textbf{0.44} & 0.10 & 0.18&\textbf{0.28} \\
GCN-MSTCN \cite{zhang2022towards}& 0.33&0.52&0.99 &0.29 &0.47 &0.85 \\
\hline
Ours & \textbf{0.07}& \underline{0.19}&0.73 & \underline{0.07} & \underline{0.17}&0.55 \\
\hline
\end{tabular}
\label{tab:oMAE}
\end{minipage}
\end{table*}

\subsection{Ablation Study}
Ablation studies, as shown in Table \ref{tab:wMAE} and Table \ref{tab:eMAE}, were conducted to assess the individual contributions of adaptive graph learning and adaptive multi-horizon learning to our method. The results validate the efficacy of our comprehensive model in anticipating surgical instruments and phases across most time horizons. Although omitting either the adaptive graph learning or multi-horizon objectives occasionally results in superior performance in some metrics, the overall performance of such ablated frameworks consistently falls short compared to our complete proposed model. This highlights that the integration of all proposed modules achieves optimal performance and may offer better generalization than models with only adaptive graph learning or multi-horizon objectives. This demonstrates the efficiency of our full model.

\subsection{Detailed Performance Analysis for $wMAE$}
For further analyzing our anticipation performance for surgical instruments and phases, we also report the results of $inMAE$ (Table \mbox{\ref{tab:inMAE}}) and $oMAE$ (Table \mbox{\ref{tab:oMAE}}), which represent in-horizon and out-of-horizon performance, respectively. Their mean corresponds to the $wMAE$ (Table~\mbox{\ref{tab:wMAE}}). Our results show significant advancements, in particular, in instrument anticipation for $inMAE$, we surpass the SOTA by 5\% in the 5-minute time horizon, and in phase anticipation, we outperform IIA-Net by 2\% in the 5-minute time horizon. Lower $inMAE$ values imply an increased precision in predicting imminent events, ensuring timely interventions during surgery. For example, during cataract surgery, accurate anticipation of the need for specific instruments allows the surgical team to prepare and respond promptly, thereby reducing delays and improving overall surgical efficiency\mbox{\cite{marafioti2021catanet}}. Additionally, our framework often ranks second-best for $oMAE$. These low $oMAE$ values demonstrate the robustness of our approach in predicting events that occur later in the surgery, which is crucial for long-term procedures where maintaining prediction accuracy over extended periods is essential. This reliability is vital for maintaining operational efficiency and patient safety.

\begin{figure}
\centering
\includegraphics[scale = 0.35]{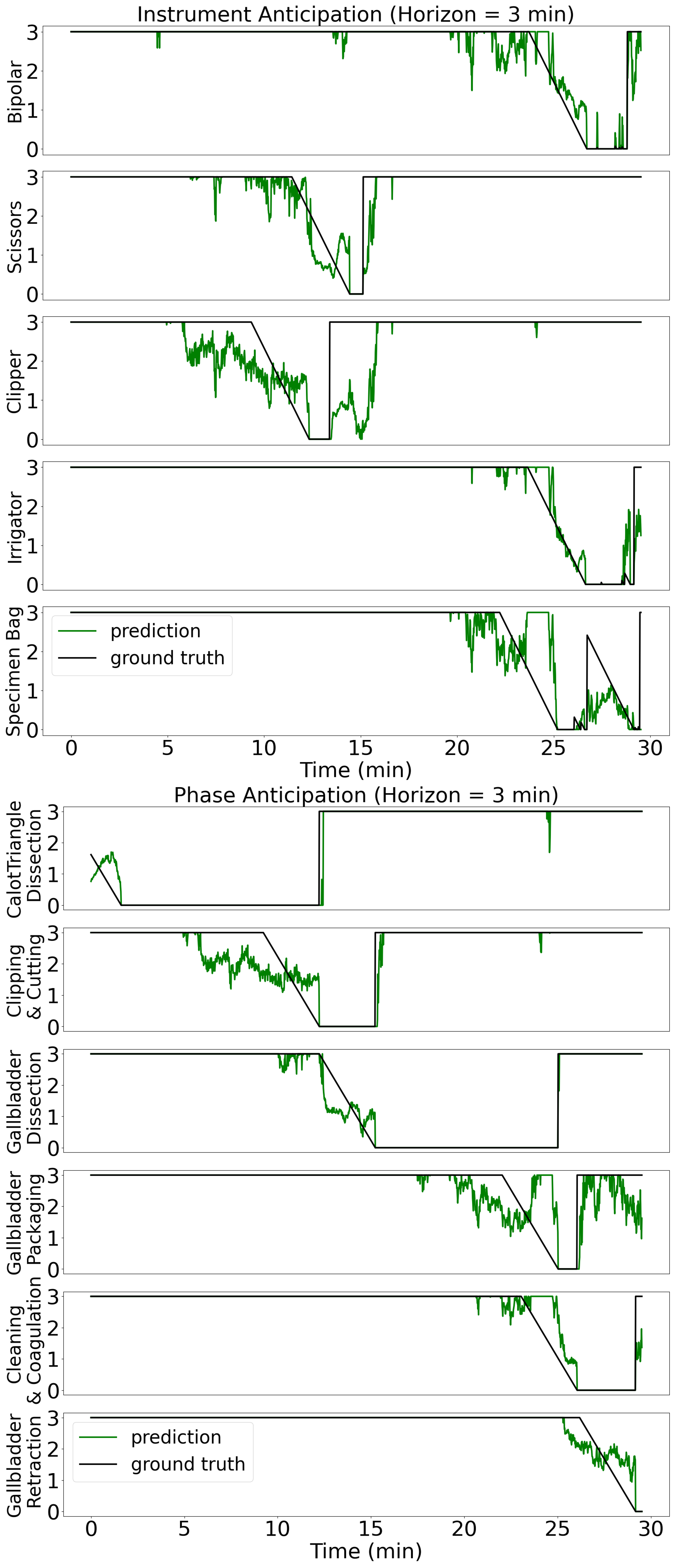}
\caption{Surgical instrument anticipation (upper section) and phase anticipation (lower section) visualization on the Cholec80 Dataset.}
\label{fig:C80}
\end{figure}
\begin{figure}
\centering
\includegraphics[scale = 0.22]{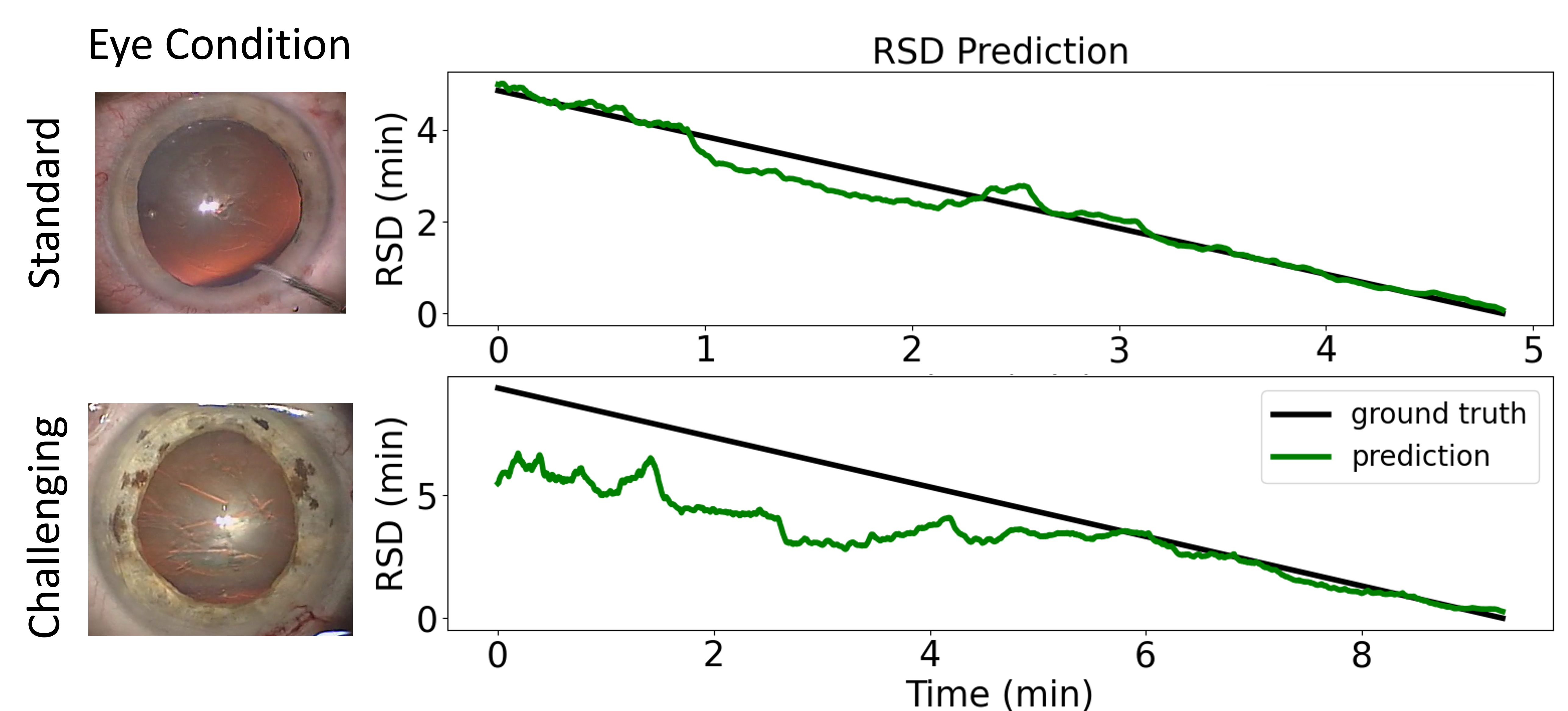}
\caption{RSD anticipation visualization on the Cataract101 dataset. Top: Standard scenario — a standard case without significant inflammation. Bottom: Challenging scenario — a challenging surgery with significant post-inflammation scarring.}
\label{fig:C101}
\end{figure}
\begin{figure}
\centering
\includegraphics[scale = 0.32]{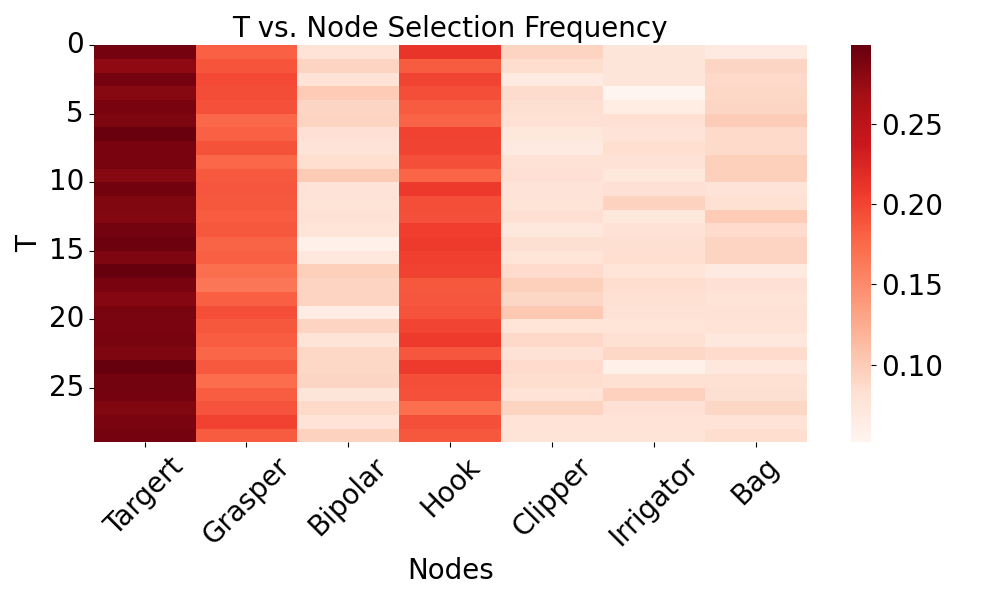}
\caption{Average importance of nodes included in optimal graph selection per minute on the Cholec80 dataset. Given that the scissors appear too infrequently to be ranked among our top 10 common candidate graphs, our candidate graphs are solely from the other 7 instruments or surgical targets.}
\label{fig:selection}
\end{figure}

\subsection{Qualitative Study}
Our qualitative study, as illustrated in Fig. \ref{fig:C80}, and Fig. \ref{fig:C101}, clearly showcases the adaptability of our system across varied surgical scenarios. 

In evaluations conducted on the Cholec80 dataset (Fig. \ref{fig:C80}), our system demonstrated reasonable performance. There is still room for improvement, especially in anticipating the use of the clipper as well as its associated clipping and cutting phases. In a cholecystectomy, the clipper is used to seal the cystic duct and artery before cutting them \cite{aspart2022clipassistnet}. The current spatial representation might not fully capture these anatomies, which could affect the performance of clipping anticipation.  In addition, occasional noise in the predictions can be observed as fluctuations in the anticipation lines. However, this does not undermine the overall reliability of our model. On the one hand, the general trend of our predictions still closely matches the ground truth for most surgical events. On the other hand, these fluctuations actually demonstrate the model’s ability to adapt its predictions in real-time to new clinical situations, such as unexpected bleeding or sudden movements of surgical instruments. In clinical applications, this continuous adjustment capability of our proposed framework enhances its reliability by allowing it to respond effectively to real-time changes in live surgical videos.

On the Cataract101 dataset, we focused on the performance difference between challenging surgery cases and standard cases. In cataract surgery, significant inflammation in the eye can lead to a challenging case\mbox{\cite{shaw2021complicated}}. Hence, we classified our test set, which consists of 20 cases, based on the observation of significant inflammation or post-inflammatory scar tissue. We found that 8 out of the 20 videos had significant inflammation and classified them as challenging cases, while 12 out of 20 were considered normal situations with no significant inflammation. The MAE for these challenging cases is 1.17, while the MAE for the standard cases is 0.88. Since the error in challenging cases is only slightly higher than the overall MAE (including both standard and challenging cases), this demonstrates that our framework can reliably anticipate both challenging and standard cases. Examples of our anticipation in both standard and challenging cases can be found in Fig. \mbox{\ref{fig:C101}}. Our system performed well with standard cases (Fig. \mbox{\ref{fig:C101}} Top) but encountered greater errors at the beginning of challenging cases, which often represent more complex surgical procedures. Nonetheless, the reduced countdown speed in difficult cases (Fig. \mbox{\ref{fig:C101}} Bottom, where dark post-inflammatory scars can be observed around the pupil) indicates our method's ability to adjust its outputs in challenging scenarios.

The visualization of graph selection is depicted in Fig. \ref{fig:selection}, with average importances calculated based on the proportion of instruments and surgical targets present in optimal graphs. The figure illustrates our graph selection continuously shifts, underscoring the dynamics of our approach. Notably, the surgical target, hook, and grasper emerge as the most frequently selected nodes, which demonstrates the ability of our graph selection to identify the core elements of a surgical procedure.

\subsection{Robustness Analysis}
\begin{figure}
    \centering
    \includegraphics[scale = 0.46]{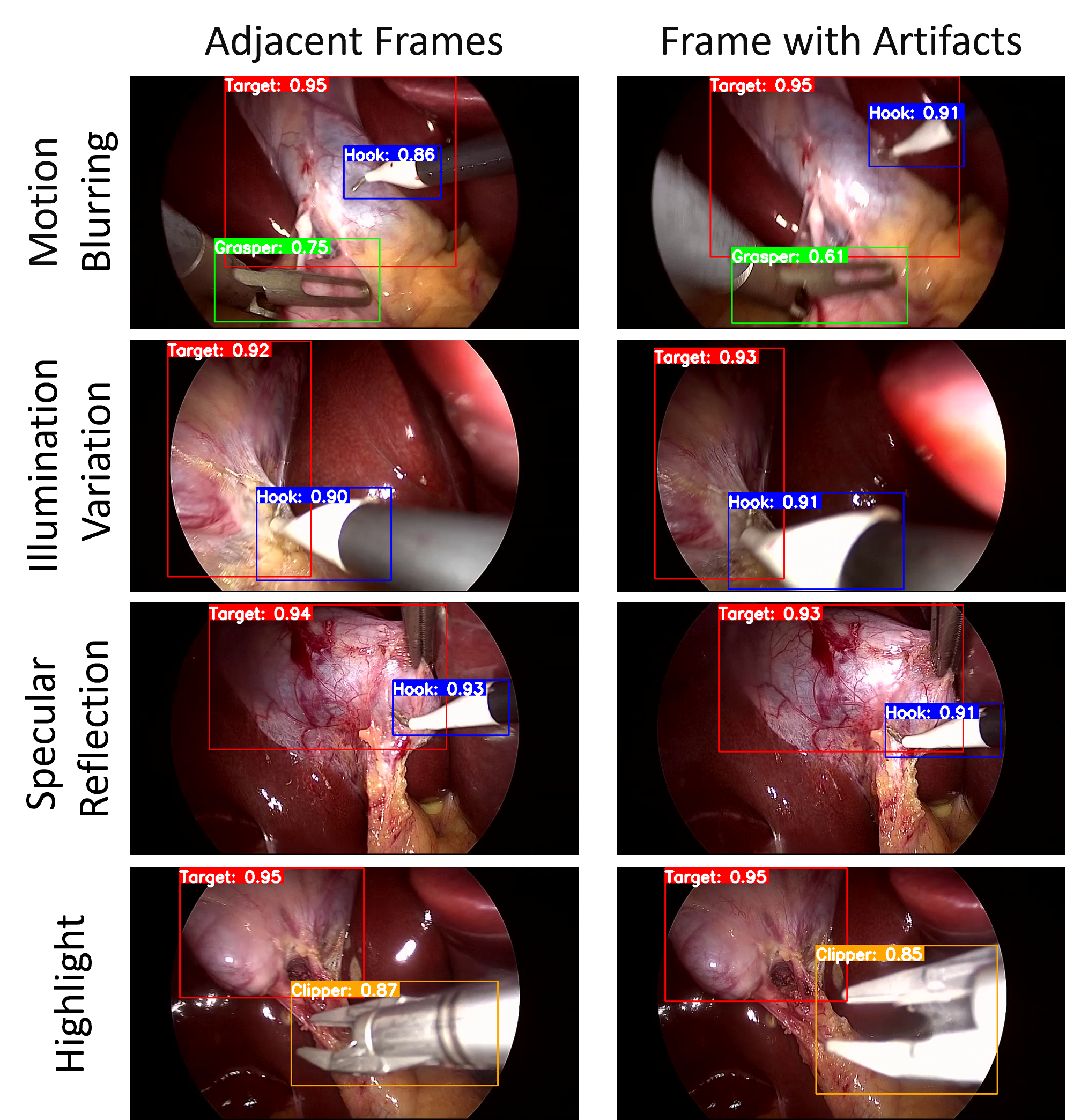}
    \caption{Bounding box performance under various common surgical imaging artifacts. Despite the presence of artifacts in frames (\textbf{Right}), the position and size of bounding boxes remain consistent with adjacent frames with fewer artifacts (\textbf{Left}), demonstrating our method's resilience to varying imaging qualities in surgical endoscopy.}
    \label{fig:artifact}
\end{figure}
\begin{figure}
\centering
\includegraphics[scale = 0.40]{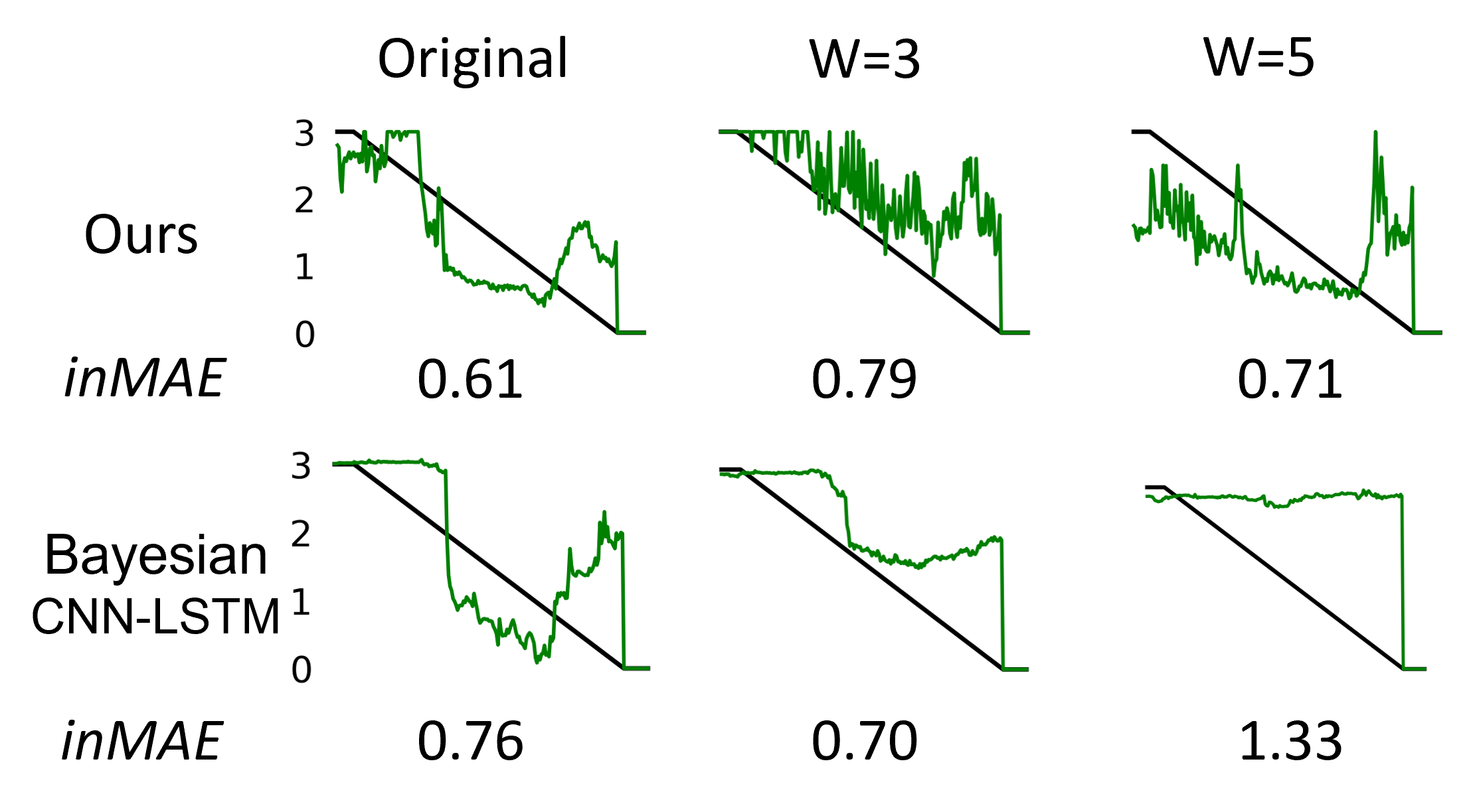}
\caption{Robustness analysis: comparing our method with visual methods in anticipating scissors under varying blur levels. Our method demonstrates better robustness across different levels of blur.}
\label{fig:robust}
\end{figure}
The visualization in Fig.~{\ref{fig:artifact}} illustrates the performance of our bounding box in handling common surgical artifacts, such as motion blurring and illumination variation\mbox{\cite{ali2021deep}}. Despite the presence of artifacts, the position and size of bounding boxes remain consistent with those in adjacent frames with fewer artifacts. This demonstrates the resilience of our method to varying imaging qualities in surgical videos.

Additionally, to demonstrate the stability of our anticipation performance, we evaluate the 3-minute instrument anticipation performance for scissors under varying degrees of blur, a frequent quality impairment in surgical videos \cite{ghamsarian2020deblurring}. Due to the undisclosed training settings of object detection models in the \ac{SOTA} method, we compared only with current pure visual methods for fairness \cite{rivoir2020rethinking}. When exposing the video to Gaussian blur windows from 1 (\emph{i.e.} original quality) to 5, we observed that, despite both spatial and visual methods showing some resilience to noise, our method consistently maintained its anticipation trend. Conversely, the anticipation trend of the visual method diminished. This underscores our hypothesis that our bounding box provides a more robust representation.

\subsection{Parameter Analysis of Adaptive Graph Learning}
\begin{figure}
\centering
\includegraphics[scale = 0.30]{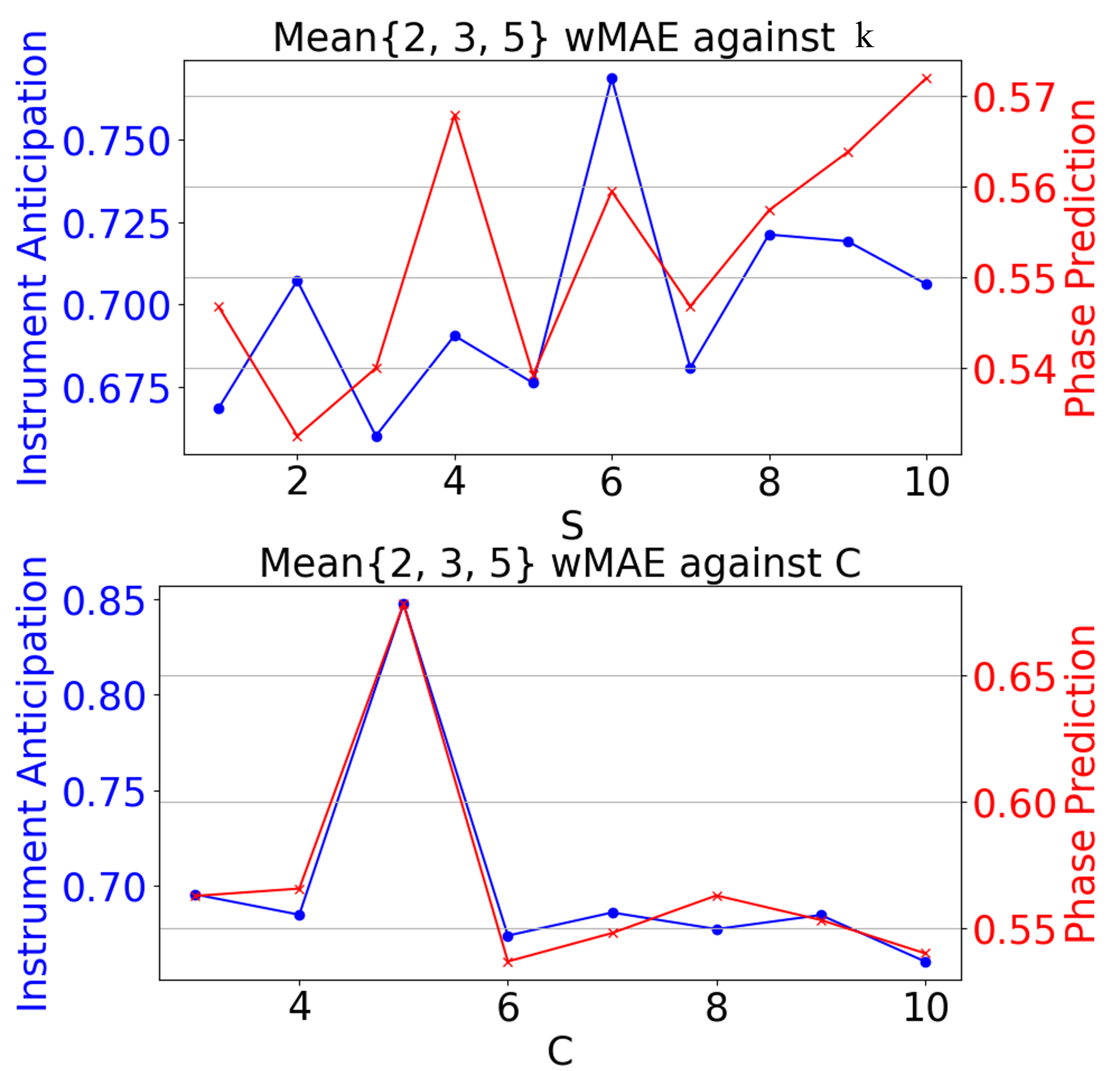}
\caption{Parameter analysis for $C$ and $S$ in our proposed framework. Our current configuration shows the optimal performance.}
\label{fig:parameter}
\end{figure}
The results of our parameter analysis for graph selections are presented in Figure \ref{fig:parameter}. Both the number of selections $k$ and the number of candidate graphs $C$ have a significant influence on the anticipation performance. A larger $C$ provides a wider array of potential surgical interactions, and $k$ determines the granularity of insights for each frame. While other configurations show a declining performance trend or an imbalance performance between two tasks, the optimal configuration for Cholec80 was $k = 3$ and $C = 10$, which yields a 0.66 and 0.54 $wMAE$ for instrument and phase anticipation respectively. This combination represents an optimal balance between the diversity of interactions and computational efficiency. While additional candidate graphs might increase model capacity, this may increase the risk of overfitting and computational demands for the \ac{RAS} system. These results illustrate the advantage of choosing fewer graph representations from a comprehensive pool of candidate graphs and the significance of the selection ratio.
\begin{table}[h]
\scriptsize
\centering
\caption{The effect of different horizon settings for training, $wMAE$ for instrument anticipation. \textbf{Bold}:the best scores.}
\begin{tabular}{l|p{1.5cm}|p{1.5cm}|p{1.5cm}}
\hline
Horizon Setting & 2 min & 3 min & 5 min\\
\hline
2 min & 0.53 & - & - \\
2, 3 min & 0.40 & 0.62 & - \\
2, 3, 5 min & 0.40 & 0.59 & 1.19\\
2, 3, 5, 7 min & \textbf{0.39} & \textbf{0.57} & \textbf{1.03} \\
2, 3, 5, 7, 9 min & 0.41 & 0.61 & 1.12\\
\hline
\end{tabular}
\label{tab:Horizon}
\end{table}
\subsection{Analysis of Training Horizon Settings}
The impact of different training horizon settings on instrument anticipation is detailed in Table \ref{tab:Horizon}. The table reveals that extending the training horizon from 2 minutes to 7 minutes consistently improves the $wMAE$ across the 2, 3, and 5-minute anticipations, achieving optimal results of 0.39, 0.57, and 1.03, respectively. Interestingly, further extending the horizon to 9 minutes slightly decreases the performance. Thus, a training horizon of 2, 3, 5, and 7 minutes emerges as the most effective setting. These results demonstrate our assumption that an appropriate multi-horizon training setting offers a global insight for each horizon to augment performance, while also suggesting the efficacy of our adaptive learning strategy.
\section{Conclusion and Discussions}
In this paper, we introduced a novel approach for surgical workflow anticipation from live video data that outperforms existing methods on a number of relevant benchmarks. Our method extracts spatial information (\emph{i.e.,} location and size) from the bounding boxes of surgical instruments and targets while also taking detection uncertainty into account. We show that this representation is particularly robust in the presence of common visual artifacts (\emph{e.g.,}~motion blur and varying lighting conditions). Additionally, we integrated this spatial information into an adaptive graph learning framework to capture dynamic interactions and demonstrate that this further improves performance. Finally, our model is trained on a novel multi-horizon objective that automatically balances short and long-term predictions and generalizes over varying surgery duration. Notably, our generalized model outperforms specialized state-of-the-art models (trained on a fixed time horizon) in multiple categories. Our empirical results highlight the advantages of our methods in short to mid-term anticipation, providing a 3\% improvement for surgical phase anticipation and 9\% for \mbox{\acl{RSD}}.

By integrating robust spatial information, adaptive graph learning, and a multi-horizon learning strategy, we provide an effective method that enables better preparation and coordination within the surgical team\mbox{\cite{rivoir2020rethinking}}, which may reduce the risk of intraoperative complications and improve patient outcomes. Additionally, the increased accuracy can lead to more efficient use of operating room resources\mbox{\cite{marafioti2021catanet}}, which provides more patients with the opportunity to receive timely treatment.

In a real-world implementation, our framework can predict the remaining duration of a live surgical procedure by continuously analyzing real-time video data and updating its predictions based on the current progress of the surgery. While there is some variability due to various factors such as patient condition, surgical plan, and anatomy, the main steps of the same type of surgery are generally consistent across different patients\mbox{\cite{garrow2021machine}}. This enables our model to dynamically adjust its predictions based on the given information, as illustrated in Figure 8, and allows it to reliably generalize over most Robotic-Assisted Surgery scenarios.

For future research, we will focus on advancing temporal feature modeling, developing a more detailed representation of surgical anatomy, and enhancing anomaly detection. First, while our model advances temporal feature modeling, its capacity for long-term dependencies can be enhanced. Future efforts might leverage diffusion-based models to strengthen long-term anticipation\mbox{\cite{chang2023design}}. Second, despite the state-of-the-art results of our current model based on bounding boxes, a more detailed representation of surgical anatomy could refine our predictions. Future iterations could integrate wider contextual spatial and visual information to more accurately reflect the surgical dynamics\mbox{\cite{qiao2022geometric}}. Third, machine learning methods generally rely on data in the training set for modeling, but surgical operations include a large variety of unseen events (\emph{e.g.,} surgical accidents) that may not be captured by our current framework. To enable the modeling and analysis of these events, future work could include anomaly detection based on energy models\mbox{\cite{han2021elsa}}. This approach enables the identification of events that have not been frequently observed and enhances the reliability of our predictions.



\section*{Acknowledgement}
This research is supported in part by the EPSRC NortHFutures project (ref: EP/X031012/1).

\bibliographystyle{IEEEtran}
\bibliography{main}

\begin{thebibliography}{10}
\providecommand{\url}[1]{#1}
\csname url@samestyle\endcsname
\providecommand{\newblock}{\relax}
\providecommand{\bibinfo}[2]{#2}
\providecommand{\BIBentrySTDinterwordspacing}{\spaceskip=0pt\relax}
\providecommand{\BIBentryALTinterwordstretchfactor}{4}
\providecommand{\BIBentryALTinterwordspacing}{\spaceskip=\fontdimen2\font plus
\BIBentryALTinterwordstretchfactor\fontdimen3\font minus \fontdimen4\font\relax}
\providecommand{\BIBforeignlanguage}[2]{{%
\expandafter\ifx\csname l@#1\endcsname\relax
\typeout{** WARNING: IEEEtran.bst: No hyphenation pattern has been}%
\typeout{** loaded for the language `#1'. Using the pattern for}%
\typeout{** the default language instead.}%
\else
\language=\csname l@#1\endcsname
\fi
#2}}
\providecommand{\BIBdecl}{\relax}
\BIBdecl

\bibitem{rivoir2020rethinking}
D.~Rivoir, S.~Bodenstedt, I.~Funke, F.~v. Bechtolsheim, M.~Distler, J.~Weitz, and S.~Speidel, ``Rethinking anticipation tasks: Uncertainty-aware anticipation of sparse surgical instrument usage for context-aware assistance,'' in \emph{International Conference on Medical Image Computing and Computer-Assisted Intervention}.\hskip 1em plus 0.5em minus 0.4em\relax Springer, 2020, pp. 752--762.

\bibitem{maier2017surgical}
L.~Maier-Hein, S.~S. Vedula, S.~Speidel, N.~Navab, R.~Kikinis, A.~Park, M.~Eisenmann, H.~Feussner, G.~Forestier, S.~Giannarou \emph{et~al.}, ``Surgical data science for next-generation interventions,'' \emph{Nature Biomedical Engineering}, vol.~1, no.~9, pp. 691--696, 2017.

\bibitem{aksamentov2017deep}
I.~Aksamentov, A.~P. Twinanda, D.~Mutter, J.~Marescaux, and N.~Padoy, ``Deep neural networks predict remaining surgery duration from cholecystectomy videos,'' in \emph{International Conference on Medical Image Computing and Computer-Assisted Intervention}.\hskip 1em plus 0.5em minus 0.4em\relax Springer, 2017, pp. 586--593.

\bibitem{twinanda2018rsdnet}
A.~P. Twinanda, G.~Yengera, D.~Mutter, J.~Marescaux, and N.~Padoy, ``Rsdnet: Learning to predict remaining surgery duration from laparoscopic videos without manual annotations,'' \emph{IEEE Transactions on Medical Imaging}, vol.~38, no.~4, pp. 1069--1078, 2018.

\bibitem{marafioti2021catanet}
A.~Marafioti, M.~Hayoz, M.~Gallardo, P.~M{\'a}rquez~Neila, S.~Wolf, M.~Zinkernagel, and R.~Sznitman, ``Catanet: Predicting remaining cataract surgery duration,'' in \emph{International Conference on Medical Image Computing and Computer-Assisted Intervention}.\hskip 1em plus 0.5em minus 0.4em\relax Springer, 2021, pp. 426--435.

\bibitem{yuan2022anticipation}
K.~Yuan, M.~Holden, S.~Gao, and W.~Lee, ``Anticipation for surgical workflow through instrument interaction and recognized signals,'' \emph{Medical Image Analysis}, vol.~82, p. 102611, 2022.

\bibitem{kirillov2023segment}
A.~Kirillov, E.~Mintun, N.~Ravi, H.~Mao, C.~Rolland, L.~Gustafson, T.~Xiao, S.~Whitehead, A.~C. Berg, W.-Y. Lo \emph{et~al.}, ``Segment anything,'' \emph{arXiv preprint arXiv:2304.02643}, 2023.

\bibitem{yuan2021surgical}
K.~Yuan, M.~Holden, S.~Gao, and W.-S. Lee, ``Surgical workflow anticipation using instrument interaction,'' in \emph{International Conference on Medical Image Computing and Computer-Assisted Intervention}.\hskip 1em plus 0.5em minus 0.4em\relax Springer, 2021, pp. 615--625.

\bibitem{wang2023context}
X.~Wang and Z.~Zhu, ``Context understanding in computer vision: A survey,'' \emph{Computer Vision and Image Understanding}, vol. 229, p. 103646, 2023.

\bibitem{li2021autonomous}
L.~Li, X.~Li, B.~Ouyang, S.~Ding, S.~Yang, and Y.~Qu, ``Autonomous multiple instruments tracking for robot-assisted laparoscopic surgery with visual tracking space vector method,'' \emph{IEEE/ASME Transactions on Mechatronics}, vol.~27, no.~2, pp. 733--743, 2021.

\bibitem{zhang2022towards}
X.~Zhang, N.~Al~Moubayed, and H.~P. Shum, ``Towards graph representation learning based surgical workflow anticipation,'' in \emph{2022 IEEE-EMBS International Conference on Biomedical and Health Informatics}.\hskip 1em plus 0.5em minus 0.4em\relax IEEE, 2022, pp. 01--04.

\bibitem{wang2022visual}
Y.~Wang, Q.~Sun, Z.~Liu, and L.~Gu, ``Visual detection and tracking algorithms for minimally invasive surgical instruments: A comprehensive review of the state-of-the-art,'' \emph{Robotics and Autonomous Systems}, vol. 149, p. 103945, 2022.

\bibitem{wu2020learning}
Y.~Wu, L.~Zhu, X.~Wang, Y.~Yang, and F.~Wu, ``Learning to anticipate egocentric actions by imagination,'' \emph{IEEE Transactions on Image Processing}, vol.~30, pp. 1143--1152, 2020.

\bibitem{sarikaya2020towards}
D.~Sarikaya and P.~Jannin, ``Towards generalizable surgical activity recognition using spatial temporal graph convolutional networks,'' \emph{arXiv preprint arXiv:2001.03728}, 2020.

\bibitem{twinanda2016endonet}
A.~P. Twinanda, S.~Shehata, D.~Mutter, J.~Marescaux, M.~De~Mathelin, and N.~Padoy, ``Endonet: a deep architecture for recognition tasks on laparoscopic videos,'' \emph{IEEE Transactions on Medical Imaging}, vol.~36, no.~1, pp. 86--97, 2016.

\bibitem{SchoeffmannTSMP18}
K.~Schoeffmann, M.~Taschwer, S.~Sarny, B.~M{\"{u}}nzer, M.~J. Primus, and D.~Putzgruber, ``Cataract-101: video dataset of 101 cataract surgeries,'' in \emph{Proceedings of the 9th {ACM} Multimedia Systems Conference}, P.~C{\'{e}}sar, M.~Zink, and N.~Murray, Eds.\hskip 1em plus 0.5em minus 0.4em\relax {ACM}, 2018, pp. 421--425.

\bibitem{bronstein2017geometric}
M.~M. Bronstein, J.~Bruna, Y.~LeCun, A.~Szlam, and P.~Vandergheynst, ``Geometric deep learning: going beyond euclidean data,'' \emph{IEEE Signal Processing Magazine}, vol.~34, no.~4, pp. 18--42, 2017.

\bibitem{qian2019review}
L.~Qian, J.~Y. Wu, S.~P. DiMaio, N.~Navab, and P.~Kazanzides, ``A review of augmented reality in robotic-assisted surgery,'' \emph{IEEE Transactions on Medical Robotics and Bionics}, vol.~2, no.~1, pp. 1--16, 2019.

\bibitem{soleymani2022surgical}
A.~Soleymani, X.~Li, and M.~Tavakoli, ``Surgical procedure understanding, evaluation, and interpretation: A dictionary factorization approach,'' \emph{IEEE Transactions on Medical Robotics and Bionics}, vol.~4, no.~2, pp. 423--435, 2022.

\bibitem{lam2022deep}
K.~Lam, F.~P.-W. Lo, Y.~An, A.~Darzi, J.~M. Kinross, S.~Purkayastha, and B.~Lo, ``Deep learning for instrument detection and assessment of operative skill in surgical videos,'' \emph{IEEE Transactions on Medical Robotics and Bionics}, vol.~4, no.~4, pp. 1068--1071, 2022.

\bibitem{liu2022towards}
Y.~Liu, Z.~Zhao, P.~Shi, and F.~Li, ``Towards surgical tools detection and operative skill assessment based on deep learning,'' \emph{IEEE Transactions on Medical Robotics and Bionics}, vol.~4, no.~1, pp. 62--71, 2022.

\bibitem{zhang2023laparoscopic}
J.~Zhang, S.~Zhou, Y.~Wang, S.~Shi, C.~Wan, H.~Zhao, X.~Cai, and H.~Ding, ``Laparoscopic image-based critical action recognition and anticipation with explainable features,'' \emph{IEEE Journal of Biomedical and Health Informatics}, 2023.

\bibitem{redmon2016you}
J.~Redmon, S.~Divvala, R.~Girshick, and A.~Farhadi, ``You only look once: Unified, real-time object detection,'' in \emph{Proceedings of the IEEE Conference on Computer Vision and Pattern Recognition}, 2016, pp. 779--788.

\bibitem{hamilton2020graph}
W.~L. Hamilton, \emph{Graph representation learning}.\hskip 1em plus 0.5em minus 0.4em\relax Morgan \& Claypool Publishers, 2020.

\bibitem{wang2019graph}
S.~Wang, Z.~Xu, C.~Yan, and J.~Huang, ``Graph convolutional nets for tool presence detection in surgical videos,'' in \emph{International Conference on Information Processing in Medical Imaging}.\hskip 1em plus 0.5em minus 0.4em\relax Springer, 2019, pp. 467--478.

\bibitem{liu2021prototypical}
J.~Liu, X.~Guo, and Y.~Yuan, ``Prototypical interaction graph for unsupervised domain adaptation in surgical instrument segmentation,'' in \emph{International Conference on Medical Image Computing and Computer Assisted Intervention}.\hskip 1em plus 0.5em minus 0.4em\relax Springer, 2021, pp. 272--281.

\bibitem{liu2021graph}
------, ``Graph-based surgical instrument adaptive segmentation via domain-common knowledge,'' \emph{IEEE Transactions on Medical Imaging}, vol.~41, no.~3, pp. 715--726, 2021.

\bibitem{islam2020learning}
M.~Islam, L.~Seenivasan, L.~C. Ming, and H.~Ren, ``Learning and reasoning with the graph structure representation in robotic surgery,'' in \emph{International Conference on Medical Image Computing and Computer-Assisted Intervention}.\hskip 1em plus 0.5em minus 0.4em\relax Springer, 2020, pp. 627--636.

\bibitem{long2021relational}
Y.~Long, J.~Y. Wu, B.~Lu, Y.~Jin, M.~Unberath, Y.-H. Liu, P.~A. Heng, and Q.~Dou, ``Relational graph learning on visual and kinematics embeddings for accurate gesture recognition in robotic surgery,'' in \emph{2021 IEEE International Conference on Robotics and Automation}.\hskip 1em plus 0.5em minus 0.4em\relax IEEE, 2021, pp. 13\,346--13\,353.

\bibitem{kadkhodamohammadi2022patg}
A.~Kadkhodamohammadi, I.~Luengo, and D.~Stoyanov, ``Patg: position-aware temporal graph networks for surgical phase recognition on laparoscopic videos,'' \emph{International Journal of Computer Assisted Radiology and Surgery}, vol.~17, no.~5, pp. 849--856, 2022.

\bibitem{kennedy2020computer}
L.~R. Kennedy-Metz, P.~Mascagni, A.~Torralba, R.~D. Dias, P.~Perona, J.~A. Shah, N.~Padoy, and M.~A. Zenati, ``Computer vision in the operating room: Opportunities and caveats,'' \emph{IEEE Transactions on Medical Robotics and Bionics}, vol.~3, no.~1, pp. 2--10, 2020.

\bibitem{shi2022recognition}
C.~Shi, Y.~Zheng, and A.~M. Fey, ``Recognition and prediction of surgical gestures and trajectories using transformer models in robot-assisted surgery,'' in \emph{2022 IEEE/RSJ International Conference on Intelligent Robots and Systems}.\hskip 1em plus 0.5em minus 0.4em\relax IEEE, 2022, pp. 8017--8024.

\bibitem{kossowsky2022predicting}
H.~Kossowsky and I.~Nisky, ``Predicting the timing of camera movements from the kinematics of instruments in robotic-assisted surgery using artificial neural networks,'' \emph{IEEE Transactions on Medical Robotics and Bionics}, vol.~4, no.~2, pp. 391--402, 2022.

\bibitem{jocher2022ultralytics}
G.~Jocher, A.~Chaurasia, A.~Stoken, J.~Borovec, Y.~Kwon, K.~Michael, J.~Fang, Z.~Yifu, C.~Wong, D.~Montes \emph{et~al.}, ``ultralytics/yolov5: v7. 0-yolov5 sota realtime instance segmentation,'' \emph{Zenodo}, 2022.

\bibitem{jin2018tool}
A.~Jin, S.~Yeung, J.~Jopling, J.~Krause, D.~Azagury, A.~Milstein, and L.~Fei-Fei, ``Tool detection and operative skill assessment in surgical videos using region-based convolutional neural networks,'' in \emph{2018 IEEE Winter Conference on Applications of Computer Vision}.\hskip 1em plus 0.5em minus 0.4em\relax IEEE, 2018, pp. 691--699.

\bibitem{FoxTS20}
M.~Fox, M.~Taschwer, and K.~Schoeffmann, ``Pixel-based tool segmentation in cataract surgery videos with mask {R-CNN},'' in \emph{33rd {IEEE} International Symposium on Computer-Based Medical Systems}.\hskip 1em plus 0.5em minus 0.4em\relax {IEEE}, 2020, pp. 565--568.

\bibitem{guo2022attention}
M.-H. Guo, T.-X. Xu, J.-J. Liu, Z.-N. Liu, P.-T. Jiang, T.-J. Mu, S.-H. Zhang, R.~R. Martin, M.-M. Cheng, and S.-M. Hu, ``Attention mechanisms in computer vision: A survey,'' \emph{Computational Visual Media}, vol.~8, no.~3, pp. 331--368, 2022.

\bibitem{mena2018learning}
G.~Mena, D.~Belanger, S.~Linderman, and J.~Snoek, ``Learning latent permutations with gumbel-sinkhorn networks,'' in \emph{International Conference on Learning Representations}, 2018.

\bibitem{kipf2017semi}
T.~N. Kipf and M.~Welling, ``Semi-supervised classification with graph convolutional networks. 2017,'' \emph{ArXiv abs/1609.02907}, 2017.

\bibitem{hu2018squeeze}
J.~Hu, L.~Shen, and G.~Sun, ``Squeeze-and-excitation networks,'' in \emph{Proceedings of the IEEE conference on Computer Vision and Pattern Recognition}, 2018, pp. 7132--7141.

\bibitem{kendall2018multi}
A.~Kendall, Y.~Gal, and R.~Cipolla, ``Multi-task learning using uncertainty to weigh losses for scene geometry and semantics,'' in \emph{Proceedings of the IEEE Conference on Computer Vision and Pattern Recognition}, 2018, pp. 7482--7491.

\bibitem{wandb}
\BIBentryALTinterwordspacing
L.~Biewald, ``Experiment tracking with weights and biases,'' 2020, software available from wandb.com. [Online]. Available: \url{https://www.wandb.com/}
\BIBentrySTDinterwordspacing

\bibitem{sener2020temporal}
F.~Sener, D.~Singhania, and A.~Yao, ``Temporal aggregate representations for long-range video understanding,'' in \emph{Computer Vision--ECCV 2020: 16th European Conference, Glasgow, UK, August 23--28, 2020, Proceedings, Part XVI 16}.\hskip 1em plus 0.5em minus 0.4em\relax Springer, 2020, pp. 154--171.

\bibitem{wu2022nonlinear}
J.~Wu, R.~Tao, and G.~Zheng, ``Nonlinear regression of remaining surgical duration via bayesian lstm-based deep negative correlation learning,'' in \emph{International Conference on Medical Image Computing and Computer-Assisted Intervention}.\hskip 1em plus 0.5em minus 0.4em\relax Springer, 2022, pp. 421--430.

\bibitem{wang2023real}
B.~Wang, L.~Li, Y.~Nakashima, R.~Kawasaki, and H.~Nagahara, ``Real-time estimation of the remaining surgery duration for cataract surgery using deep convolutional neural networks and long short-term memory,'' \emph{BMC Medical Informatics and Decision Making}, vol.~23, no.~1, p.~80, 2023.

\bibitem{copenhaver2017improving}
M.~S. Copenhaver, T.~H. Friend, C.~Fitzgerald-Brown, M.~Fernandez, M.~Addesa, J.~Cassidy, M.~Rosa, J.~Ouellette, J.~Plunkett, D.~Spracklin \emph{et~al.}, ``Improving operating room and surgical instrumentation efficiency, safety, and communication via the implementation of emergency laparoscopic cholecystectomy and appendectomy conversion case carts,'' \emph{Perioperative Care and Operating Room Management}, vol.~8, pp. 33--37, 2017.

\bibitem{bogdanovic2015adaptive}
J.~Bogdanovic, J.~Perry, M.~Guggenheim, and T.~Manser, ``Adaptive coordination in surgical teams: an interview study,'' \emph{BMC health services research}, vol.~15, pp. 1--12, 2015.

\bibitem{aspart2022clipassistnet}
F.~Aspart, J.~L. Bolmgren, J.~L. Lavanchy, G.~Beldi, M.~S. Woods, N.~Padoy, and E.~Hosgor, ``Clipassistnet: bringing real-time safety feedback to operating rooms,'' \emph{International Journal of Computer Assisted Radiology and Surgery}, vol.~17, pp. 5--13, 2022.

\bibitem{shaw2021complicated}
E.~Shaw and B.~C. Patel, ``Complicated cataract,'' 2021.

\bibitem{ali2021deep}
S.~Ali, F.~Zhou, A.~Bailey, B.~Braden, J.~E. East, X.~Lu, and J.~Rittscher, ``A deep learning framework for quality assessment and restoration in video endoscopy,'' \emph{Medical image analysis}, vol.~68, p. 101900, 2021.

\bibitem{ghamsarian2020deblurring}
N.~Ghamsarian, M.~Taschwer, and K.~Schoeffmann, ``Deblurring cataract surgery videos using a multi-scale deconvolutional neural network,'' in \emph{2020 IEEE 17th International Symposium on Biomedical Imaging}.\hskip 1em plus 0.5em minus 0.4em\relax IEEE, 2020, pp. 872--876.

\bibitem{garrow2021machine}
C.~R. Garrow, K.-F. Kowalewski, L.~Li, M.~Wagner, M.~W. Schmidt, S.~Engelhardt, D.~A. Hashimoto, H.~G. Kenngott, S.~Bodenstedt, S.~Speidel \emph{et~al.}, ``Machine learning for surgical phase recognition: a systematic review,'' \emph{Annals of surgery}, vol. 273, no.~4, pp. 684--693, 2021.

\bibitem{chang2023design}
Z.~Chang, G.~A. Koulieris, and H.~P. Shum, ``On the design fundamentals of diffusion models: A survey,'' \emph{arXiv preprint arXiv:2306.04542}, 2023.

\bibitem{qiao2022geometric}
T.~Qiao, Q.~Men, F.~W. Li, Y.~Kubotani, S.~Morishima, and H.~P. Shum, ``Geometric features informed multi-person human-object interaction recognition in videos,'' in \emph{European Conference on Computer Vision}.\hskip 1em plus 0.5em minus 0.4em\relax Springer, 2022, pp. 474--491.

\bibitem{han2021elsa}
S.~Han, H.~Song, S.~Lee, S.~Park, and M.~Cha, ``Elsa: Energy-based learning for semi-supervised anomaly detection,'' in \emph{The 32nd British Machine Vision Conference}.\hskip 1em plus 0.5em minus 0.4em\relax British Machine Vision Association, 2021.

\end{thebibliography}

\begin{IEEEbiography}
[{\includegraphics[width=1in,height=1.25in,clip,keepaspectratio]{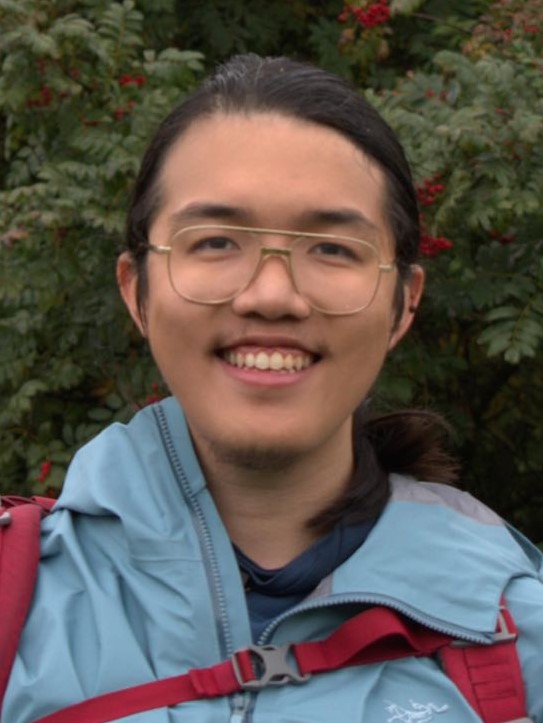}}]{Francis Xiatian Zhang} 
(Student Member, IEEE) is a PhD student in Computer Science at Durham University, researching on biomedical engineering and health informatics. Before this, he received his MRes degree from King's College London, his MSc degree from the University of Southampton, and his bachelor's degree from Beijing University of Chinese Medicine. Currently, he serves as a research assistant at Durham University and previously held a similar position at Northumbria University.
\end{IEEEbiography}
\begin{IEEEbiography}[{\includegraphics[width=1in,height=1.25in,clip,keepaspectratio]{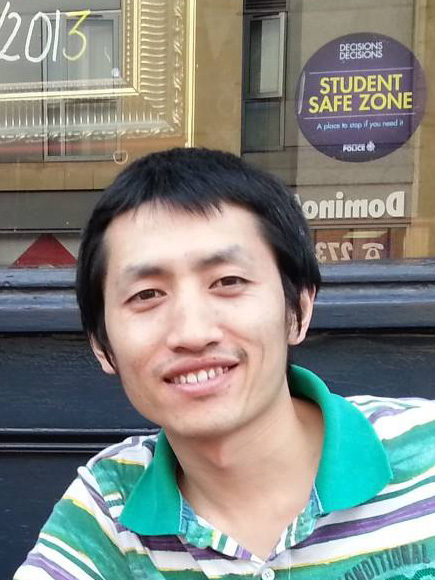}}]{Jingjing Deng}
(Member, IEEE) received the M.Sc. by Research and Ph.D. degrees in computer science from Swansea University, UK, in 2012 and 2017 respectively. He is currently an Assistant Professor in the Department of Computer Science, Durham University, UK. His research interests are computational and mathematical intelligence, non-linear analysis and computing, and applications in biomedical and physical sciences.
\end{IEEEbiography}
\begin{IEEEbiography}
[{\includegraphics[width=1in,height=1.25in,clip,keepaspectratio]{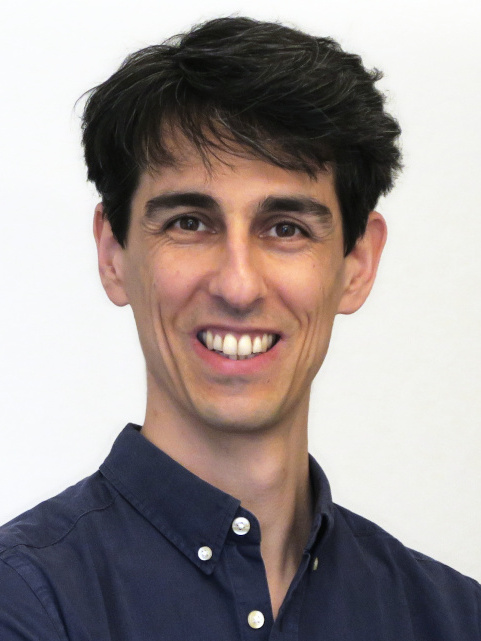}}]{Robert Lieck}
is an Assistant Professor in Computer Science at Durham University. He has a background in physics and philosophy and a PhD in machine learning and robotics from the University of Stuttgart, Germany. His research is in the area of machine learning and artificial intelligence with a focus on complex temporal structures, reinforcement learning, deep learning, and differentiable programming. He has more than a decade of experience in interdisciplinary research and applications, including ethics in AI, musical structure and perception, cognitive modeling, and medical image processing.
\end{IEEEbiography}
\begin{IEEEbiography}
[{\includegraphics[width=1in,height=1.25in,clip,keepaspectratio]{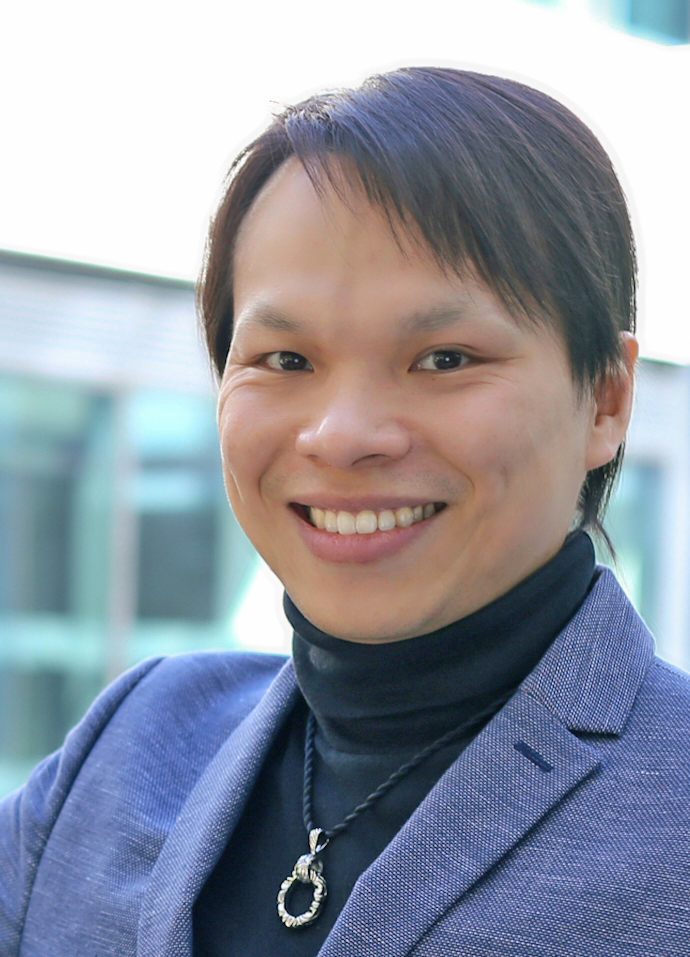}}]{Hubert P. H. Shum}
(Senior Member, IEEE) is a Professor of Visual Computing and the Director of Research of the Department of Computer Science at Durham University, specialising in Spatio-Temporal Modelling and Responsible AI. He is also a Co-Founder and the Co-Director of Durham University Space Research Centre. Before this, he was an Associate Professor at Northumbria University and a Postdoctoral Researcher at RIKEN Japan. He received his PhD degree from the University of Edinburgh. He chaired conferences such as Pacific Graphics, BMVC and SCA, and has authored over 180 research publications.
\end{IEEEbiography}
\end{document}